\documentclass[letterpaper, 10 pt, conference]{ieeeconf}  

\IEEEoverridecommandlockouts                              

\usepackage{amsopn}

\newcommand{\bb}{\mathbf{b}}

\input{gnuplot}

\renewcommand{\textfloatsep}{0em}

\usepackage{graphics} 
\usepackage{epsfig} 
\usepackage{times} 
\usepackage{amsmath} 
\usepackage{amssymb}  

\makeatletter
\let\NAT@parse\undefined
\makeatother

\usepackage[colorlinks=true,linkcolor=blue,citecolor=blue,urlcolor=blue,pagebackref]{hyperref}
\usepackage[font=footnotesize]{caption}
\usepackage[font=footnotesize]{subcaption}
\usepackage{adjustbox}
\usepackage{cite}

\usepackage{glossaries-extra}
\setabbreviationstyle[acronym]{long-short}
\glssetcategoryattribute{acronym}{nohyperfirst}{true}

\makeglossaries

\newacronym{slam}{SLAM}{Simultaneous Localization and Mapping}
\newacronym{vo}{VO}{Visual Odometry}
\newacronym{vpr}{VPR}{Visual Place Recognition}
\newacronym{kc}{KC}{Keypoint Coordinates}
\newacronym{sl}{SL}{Semantic Labels}
\newacronym{pr}{PR}{Precision-Recall}
\newacronym{bof}{BOF}{Bag-of-Features}
\newacronym{lsh}{LSH}{Locality Sensitive Hashing}
\newacronym{hbst}{HBST}{Hamming Binary Search Tree}
\newacronym{bst}{BST}{Binary Search Tree}
\newacronym{bf}{BF}{Brute-Force}
\newacronym{map}{mAP}{Mean Average Precision}
\newacronym{roc}{ROC}{Receiver Operating Characteristic}
\newacronym{dbow}{DBoW}{Bags of Binary Words}

\def\secref#1{Sec.~\ref{#1}}
\def\figref#1{Fig.~\ref{#1}}
\def\tabref#1{Tab.~\ref{#1}}
\def\eqref#1{Eq.~(\ref{#1})}

\def\etal{~\emph{et al.}}

\def\descriptor{\mathbf{d}}
\def\augmentationweight{\lambda}

\title{\LARGE \bf Adding Cues to Binary Feature Descriptors for Visual Place Recognition}
\author{Dominik Schlegel and Giorgio Grisetti
\thanks{Both authors are with the Department of Computer, Control and Management Engineering,
        Sapienza University of Rome, Rome, Italy
        {\tt \{lastname\}@diag.uniroma1.it}}%
}

\begin{document}

\maketitle
\thispagestyle{empty}
\pagestyle{empty}

\begin{abstract}
  In this paper we propose an approach to embed continuous and
  selector cues in binary feature descriptors used for visual place
  recognition. The embedding is achieved by extending each feature
  descriptor with a binary string that encodes a cue and supports the 
  Hamming distance metric. Augmenting the descriptors in such a way has the advantage of being
  transparent to the procedure used to compare them.
  We present two concrete applications of our methodology,
  demonstrating the two considered types of cues. In addition to that, we conducted on these
  applications a broad quantitative and comparative evaluation covering five
  benchmark datasets and several state-of-the-art image retrieval
  approaches in combination with various binary descriptor types.
\end{abstract}

\section{Introduction}\label{sec:introduction}

\gls{vpr} has been receiving increased attention over the last
decade~\cite{2003-torralba-context, 2012-galvez-bow, 2016-lowry-place-recognition-survey, 2017-zhu-place-recognition}.
The task of a \gls{vpr} system is the retrieval (i.e. recognition) of similar images
from a database of visited places that match a current image.
This \emph{similarity search} problem is typically addressed through image comparison.
To reduce the dimensionality of the problem and to gain robusteness to
viewpoint and illumination variations, it is common to represent an
image by a set of \emph{feature descriptors}~\cite{2001-oliva-gist, 2004-lowe-sift, 2005-dalal-hog,
2010-calonder-brief, 2011-rublee-orb, 2011-leutenegger-brisk, 2011-alcantarilla-fast, 
2012-alahi-freak, 2012-strecha-ldahash, 2013-trzcinski-boosting, 2015-balntas-bold}.
These descriptors are generally represented by floating-point or binary \emph{vectors}
that admit a certain distance metric.
Comparing two images is then reduced to measuring the cumulative
distance between their corresponding descriptors.

Typically, descriptors are computed exclusively from image pixel data.
On one hand, this allows to use these descriptors on any system that needs \gls{vpr} capabilities.
On the other hand, if external place-specific \emph{cues} (e.g. GPS coordinates) are available,
the \gls{vpr} system based on descriptors cannot immediately benefit from that additional information.
The cues can be incorporated in a pre- or post-filtering stage.
In general however, this strategy requires a major modification of the similarity search approach~\cite{2007-lv-lsh, 2008-jegou-hamming, 2016-sankaran-parameter}.

\begin{figure}[t!]
  \vspace{5pt}
  \centering
  \begin{subfigure}[]{\columnwidth}
    \includegraphics[width=\columnwidth]{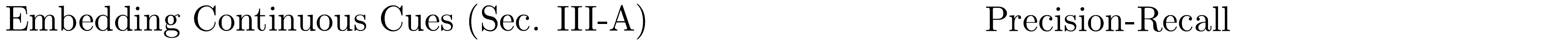}
    \raisebox{12pt}{\adjustbox{width=0.6\columnwidth, trim=94pt 0pt 0pt 0pt, clip}{\includegraphics[width=\columnwidth]{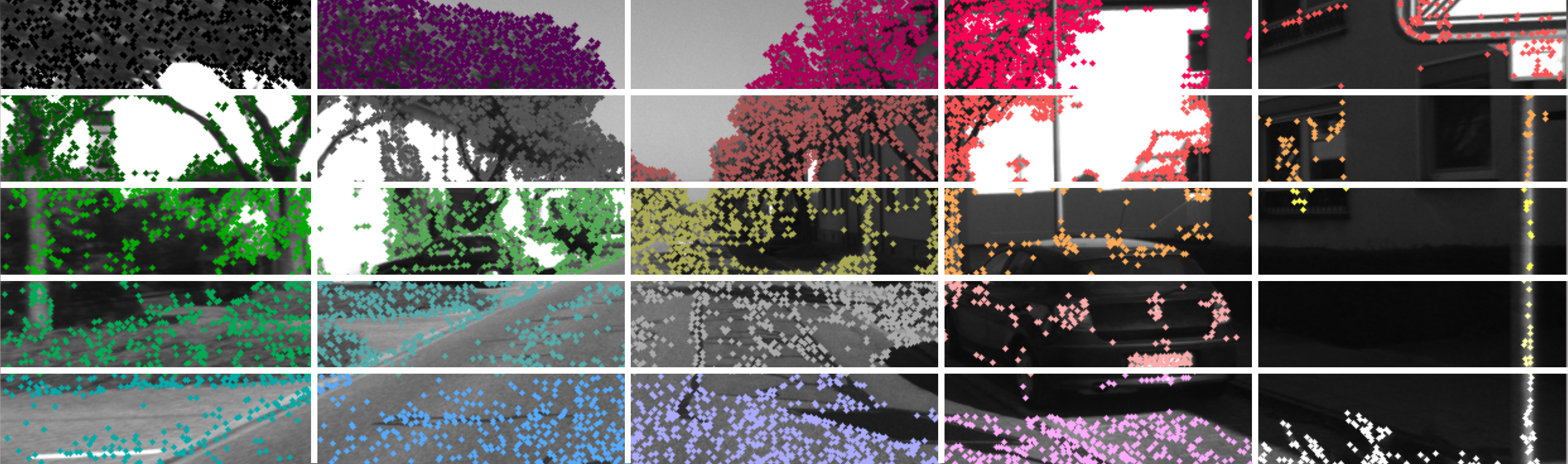}}}
    \adjustbox{width=0.39\columnwidth, trim=40pt 0pt 50pt 5pt, clip}{
    \begin{picture}(7200.00,5040.00)%
    \gdef\gplbacktext{}
    \gdef\gplfronttext{}
    \gplgaddtomacro\gplbacktext{%
      \csname LTb\endcsname%
      \put(1025,484){\makebox(0,0){\strut{}$0.5$}}%
      \csname LTb\endcsname%
      \put(3061,484){\makebox(0,0){\strut{}$0.75$}}%
      \csname LTb\endcsname%
      \put(5096,484){\makebox(0,0){\strut{}$1$}}%
      \csname LTb\endcsname%
      \put(5228,704){\makebox(0,0)[l]{\strut{}$0$}}%
      \csname LTb\endcsname%
      \put(5228,1722){\makebox(0,0)[l]{\strut{}$0.25$}}%
      \csname LTb\endcsname%
      \put(5228,2740){\makebox(0,0)[l]{\strut{}$0.5$}}%
      \csname LTb\endcsname%
      \put(5228,3757){\makebox(0,0)[l]{\strut{}$0.75$}}%
      \csname LTb\endcsname%
      \put(5228,4775){\makebox(0,0)[l]{\strut{}$1$}}%
    }%
    \gplgaddtomacro\gplfronttext{%
      \csname LTb\endcsname%
      \put(5997,2739){\rotatebox{-270}{\makebox(0,0){\strut{}Precision (correct/reported associations)}}}%
      \put(3060,154){\makebox(0,0){\strut{}Recall (correct/possible associations)}}%
      \csname LTb\endcsname%
      \put(2345,2417){\makebox(0,0)[r]{\strut{}BF, $\lambda=0$}}%
      \csname LTb\endcsname%
      \put(2345,2197){\makebox(0,0)[r]{\strut{}BF, $\lambda=16$}}%
      \csname LTb\endcsname%
      \put(2345,1977){\makebox(0,0)[r]{\strut{}LSH, $\lambda=0$}}%
      \csname LTb\endcsname%
      \put(2345,1757){\makebox(0,0)[r]{\strut{}LSH, $\lambda=16$}}%
      \csname LTb\endcsname%
      \put(2345,1537){\makebox(0,0)[r]{\strut{}BOF, $\lambda=0$}}%
      \csname LTb\endcsname%
      \put(2345,1317){\makebox(0,0)[r]{\strut{}BOF, $\lambda=16$}}%
      \csname LTb\endcsname%
      \put(2345,1097){\makebox(0,0)[r]{\strut{}BST, $\lambda=0$}}%
      \csname LTb\endcsname%
      \put(2345,877){\makebox(0,0)[r]{\strut{}BST, $\lambda=16$}}%
    }%
    \gplbacktext
    \put(0,0){\includegraphics{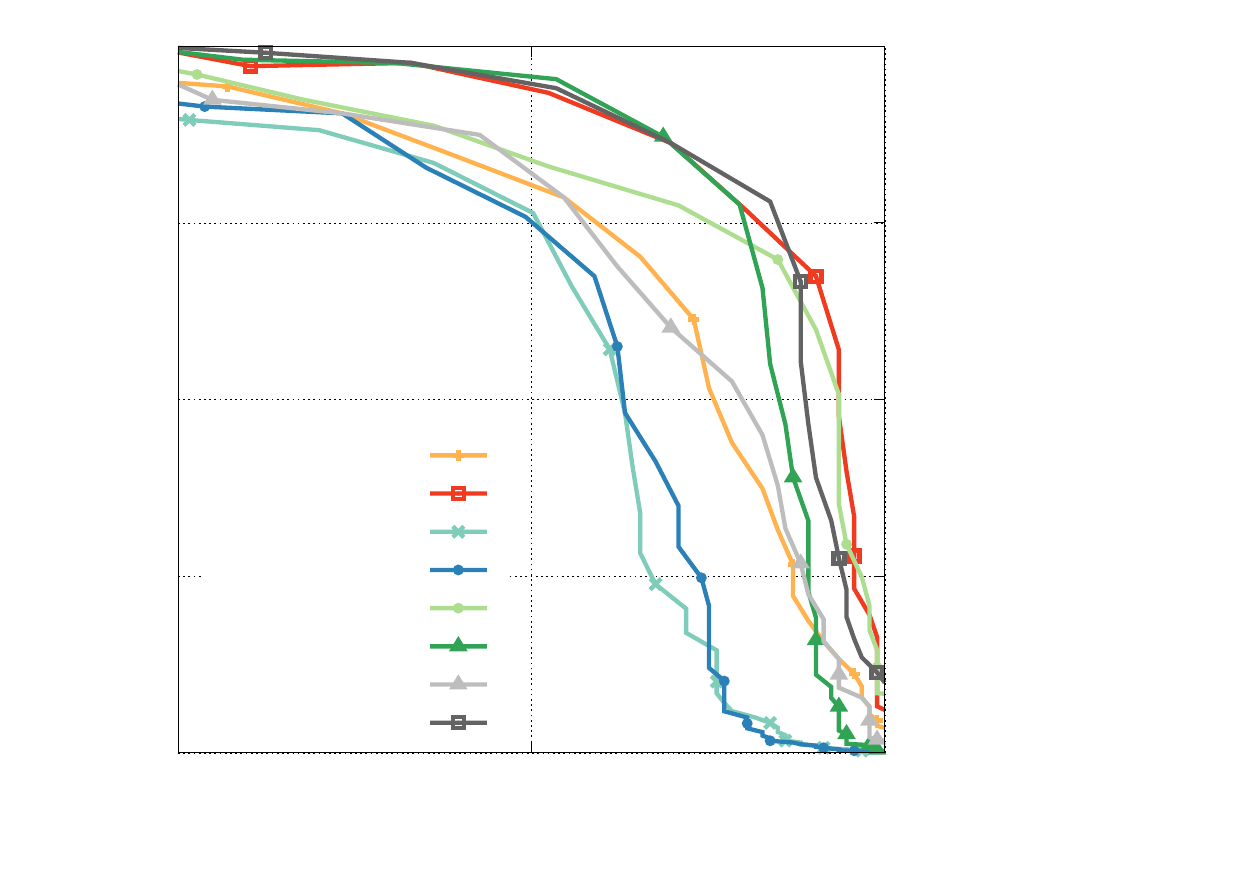}}%
    \gplfronttext
  \end{picture}
}
  \end{subfigure}
  \begin{subfigure}[]{\columnwidth}
    \includegraphics[width=\columnwidth]{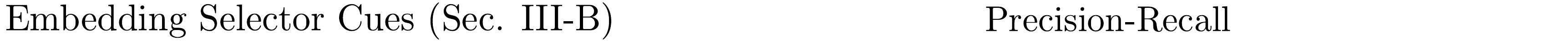}
    \raisebox{12pt}{\adjustbox{width=0.6\columnwidth, trim=94pt 0pt 0pt 0pt, clip}{\includegraphics[width=\columnwidth]{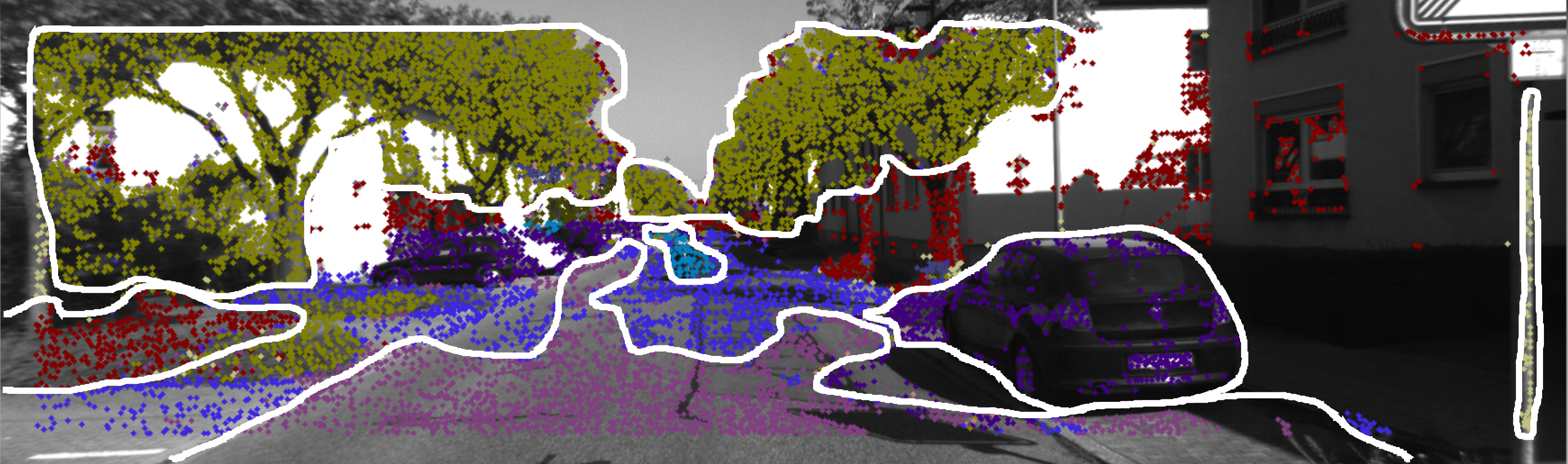}}}
    \adjustbox{width=0.39\columnwidth, trim=40pt 0pt 50pt 5pt, clip}{
    \begin{picture}(7200.00,5040.00)%
    \gdef\gplbacktext{}
    \gdef\gplfronttext{}
    \gplgaddtomacro\gplbacktext{%
      \csname LTb\endcsname%
      \put(1025,484){\makebox(0,0){\strut{}$0.5$}}%
      \csname LTb\endcsname%
      \put(3061,484){\makebox(0,0){\strut{}$0.75$}}%
      \csname LTb\endcsname%
      \put(5096,484){\makebox(0,0){\strut{}$1$}}%
      \csname LTb\endcsname%
      \put(5228,704){\makebox(0,0)[l]{\strut{}$0$}}%
      \csname LTb\endcsname%
      \put(5228,1722){\makebox(0,0)[l]{\strut{}$0.25$}}%
      \csname LTb\endcsname%
      \put(5228,2740){\makebox(0,0)[l]{\strut{}$0.5$}}%
      \csname LTb\endcsname%
      \put(5228,3757){\makebox(0,0)[l]{\strut{}$0.75$}}%
      \csname LTb\endcsname%
      \put(5228,4775){\makebox(0,0)[l]{\strut{}$1$}}%
    }%
    \gplgaddtomacro\gplfronttext{%
      \csname LTb\endcsname%
      \put(5997,2739){\rotatebox{-270}{\makebox(0,0){\strut{}Precision (correct/reported associations)}}}%
      \put(3060,154){\makebox(0,0){\strut{}Recall (correct/possible associations)}}%
      \csname LTb\endcsname%
      \put(2213,2417){\makebox(0,0)[r]{\strut{}BF, $\lambda=0$}}%
      \csname LTb\endcsname%
      \put(2213,2197){\makebox(0,0)[r]{\strut{}BF, $\lambda=2$}}%
      \csname LTb\endcsname%
      \put(2213,1977){\makebox(0,0)[r]{\strut{}LSH, $\lambda=0$}}%
      \csname LTb\endcsname%
      \put(2213,1757){\makebox(0,0)[r]{\strut{}LSH, $\lambda=2$}}%
      \csname LTb\endcsname%
      \put(2213,1537){\makebox(0,0)[r]{\strut{}BOF, $\lambda=0$}}%
      \csname LTb\endcsname%
      \put(2213,1317){\makebox(0,0)[r]{\strut{}BOF, $\lambda=2$}}%
      \csname LTb\endcsname%
      \put(2213,1097){\makebox(0,0)[r]{\strut{}BST, $\lambda=0$}}%
      \csname LTb\endcsname%
      \put(2213,877){\makebox(0,0)[r]{\strut{}BST, $\lambda=2$}}%
    }%
    \gplbacktext
    \put(0,0){\includegraphics{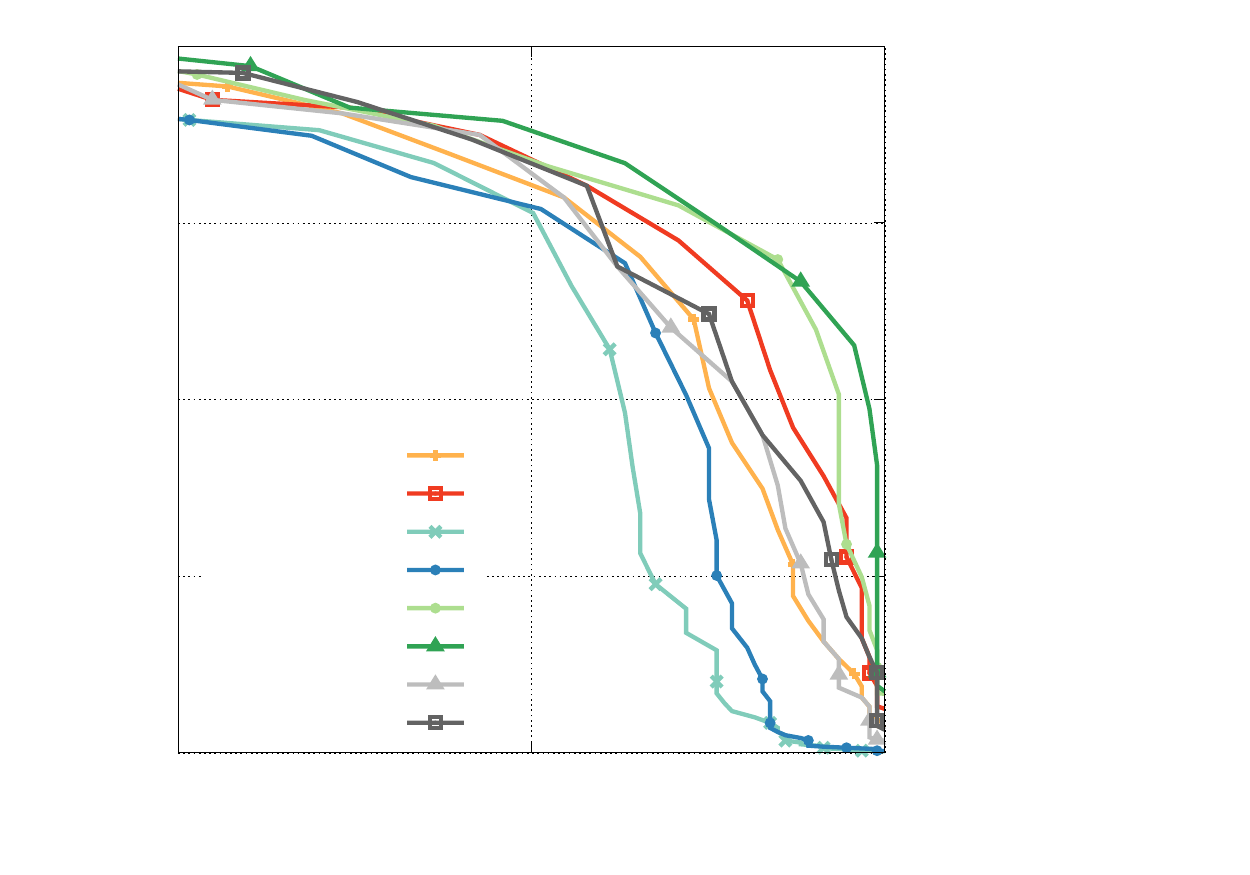}}%
    \gplfronttext
  \end{picture}
}
  \end{subfigure}
  \vspace{-0pt}
  \caption{Image extracts with FAST keypoints and BRIEF-256 descriptors,
           accompanied by a \gls{pr} evaluation.
           Top: Keypoint coordinates as additional cues.
           Bottom: Semantic labels as additional cues.}
  \label{fig:motivation}
  \vspace{0pt}
\end{figure}
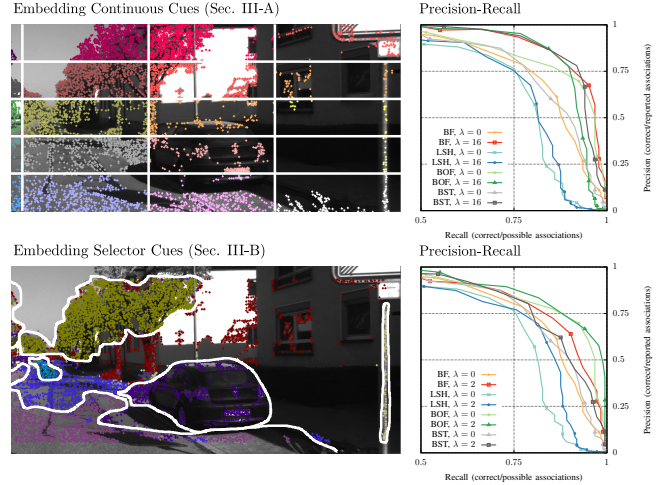

Alternatively, one can \emph{embed} the cues as additional dimensions in the descriptor vector,
without modifying the search approach.
During the search, the additional cues will contribute to the distance metric,
resulting in small cumulative distances when the described image portions \emph{and} the cues are
similar~\cite{2007-mikolajczyk-improving, 2010-harada-improving, 2013-kottman-improving, 2014-wang-cs, 2016-xiao-bag}.

Embedding continuous cues is straightforward when the descriptors are floating-point vectors.
Since in this case the common distance metric is the $L_2$-norm.
Conversely, \emph{binary descriptors} are compared with the Hamming distance $L_\mathcal{H}$~\cite{1950-hamming},
that is the number of mismatching bits between the compared descriptors.
A \emph{binary cue} can be added to a binary descriptor, analogously to the floating-point case.
A continuous cue, however, cannot be added straightforwardly to a binary descriptor
since the Hamming distance between two floating-point values does not reflect their arithmetic difference, unless they are equal.
Thus the binary representation of a
floating-point value does not qualify as a valid binary feature for
the cue that can be used to augment the descriptor.

In this paper, we address the problem of adding \emph{continuous} and
\emph{selector cues} to binary descriptors.
We propose an approach to compute binary features from continuous cues
which can be used to augment the descriptors. These binary strings support
the Hamming distance, thus comparison approaches that are based on
this norm do not have to be changed to deal with the augmented
descriptors. Additionally, we elaborate on how to encode a
selector cue. Selector cues assume a specific value from a finite set
and the distance between two such cues is either 0 if
they are the same or 1 otherwise.
A typical type of selector is the semantic class of a pixel returned by a classifier.

We demonstrate the applicability of our approach for two example cues
and verify its benefit for \gls{vpr} in an extensive experimental evaluation
on five public benchmark datasets with several state-of-the-art \gls{vpr} approaches.
In \figref{fig:motivation} we display a comparative performance evaluation
of our example cues on the KITTI \gls{slam} dataset.
All results presented in this paper can be reproduced using our open-source C++
implementation\footnote{Source at: \url{www.gitlab.com/srrg-software/srrg_bench}}.

\section{Related Work}\label{sec:related-work}

Many efforts have been made to improve existing, handcrafted local binary descriptors,
creating the broad variety available today~\cite{2016-madeo-fast}.
Years before the introduction of the well-known BRIEF binary descriptor~\cite{2010-calonder-brief},
Mikolajczyk and Matas~\cite{2007-mikolajczyk-improving} presented an approach to improve
the state-of-the-art SIFT~\cite{2004-lowe-sift} floating point descriptor for matching.
In their work, they transformed the original descriptors
based on learned optimal projections under the Mahalanobis distance.
Another early work about enhancing scene classification and object recognition
for various global and local descriptors has been reported by Harada\etal~\cite{2010-harada-improving}.
Harada\etal~showed that by consideration of locality, correlation and the concatenation of local descriptors
the matching accuracy can be raised significantly.
None of these works considered the local binary descriptors
which we evaluate in this paper.

Shortly after the release of the pioneering BRIEF binary descriptor,
the well-known ORB~\cite{2011-rublee-orb} and BRISK~\cite{2011-leutenegger-brisk} descriptors were introduced.
ORB improved BRIEF by adding rotation and scale invariance,
while BRISK improved BRIEF by considering a Gaussian pixel average for the descriptor computation.
After the introduction of the bio-inspired FREAK binary descriptor~\cite{2012-alahi-freak},
Kottman~\cite{2013-kottman-improving} presented with LFREAK an enhanced version of FREAK by including
the spatial arrangement between multiple keypoints.
Similarly, Wang\etal~\cite{2014-wang-cs} proposed with CS-FREAK a FREAK variant
that considers the neighborhood intensity information of its surrounding sampling points.
Recently, Xiao\etal~\cite{2016-xiao-bag} introduced a binary descriptor (BAG) suitable for RGB-D image processing.
Using binary tests, Xiao\etal~add geometric cues based on point depth
to the local binary pattern based on image intensity in a patch.
In contrast to our work, their cues are not computed additively to the descriptor
and neither transfer quantifiable cue distances.

For the remainder of this article, if not specified otherwise,
with \emph{descriptors} we always refer to local binary descriptors.
Binary descriptors are generally compared by their Hamming distance.
Sankaran\etal~\cite{2016-sankaran-parameter} investigated the effects of a weighted Hamming distance
and a thresholded binary testing on ORB.
Sankaran\etal~show that an improved accuracy can be achieved on various datasets.

Often, these descriptor improvements are tailored to a specific search approach for a specific descriptor type.
Among the most popular similarity search approaches is the \gls{bof} approach of Sivic and Zisserman~\cite{2003-sivic-bof}.
\gls{bof} reduces the high dimensionality of the search problem
by quantization of descriptors into \emph{visual words},
for which a frequency histogram can be obtained that describes an entire image.
Subsequently, Jegou\etal~\cite{2008-jegou-hamming} presented a twofold improvement of the \gls{bof} approach.
By differentiation on the position of a local descriptor within its k-means cluster,
and adding a consistency check on the keypoint's angle and scale,
precision and runtime of \gls{bof} could be improved significantly.
The euclidean descriptor positions are encoded into binary signatures,
with a procedure introduced as Hamming Embedding.
In contrast to their procedure, which \emph{converts descriptors into binary strings},
our approach generates binary strings based on \emph{auxiliary information}
from an arbitrary source, that are \emph{added} to a binary descriptor.
Furthermore, our approach does not require a learning phase
and is completely independent of the target similarity search approach. 
In their \gls{dbow} open-source library,
G{\'a}lvez-L{\'o}pez and Tard{\'o}s~\cite{2012-galvez-bow} implemented
the \gls{bof} approach and added further improvements while
making it accessible for the \gls{slam} research community.

Multi-probe \gls{lsh} by Lv\etal~\cite{2007-lv-lsh} is a popular similarity search approach,
that reduces the dimensionality of the search problem with hashing.
Depending on the chosen hash key lengths and the number of hashing tables,
\gls{lsh} requires significantly more memory than \gls{bof} to index descriptors in its database.

In \gls{hbst}, one of our works~\cite{2018-schlegel-hbst},
we presented a \gls{bst} approach for fast binary descriptor search.
By arranging the descriptors in a tree based on particular bits,
the search problem dimensionality is reduced by one
and the number of search candidates is halved, for each traversed node.
\gls{hbst} achieves excelling search speed,
while maintaining a meaningful accuracy in small and large scale scenarios.

Note that the mentioned similarity search approaches~\cite{2003-sivic-bof, 2007-lv-lsh, 2018-schlegel-hbst}
are only guaranteed to find an \emph{approximate nearest neighbor},
as opposed to the exhaustive \gls{bf} search,
which always returns the best match.

The aim of this paper is to present an approach that 
improves the \gls{vpr} image retrieval and descriptor matching precision
of an existing similarity search system by considering additional, place-relevant cues.
While most of the discussed works
require a modification of the descriptor computation or similarity search method,
our approach is purely supplementary.
Additionally, our approach has a negligible memory footprint and
comes at a vanishing computational cost.
Since we exploit the particular conditions of \gls{vpr} to our advantage,
our approach is solely targeted at \gls{vpr}.

\section{Our Approach}\label{sec:our-approach}

In this paper we address image retrieval based on feature descriptor matching.
More specifically, to obtain a measure for the similarity between images,
we compare their corresponding descriptors.

Feature descriptors are vectors that encode the local appearance of
an image around a point of interest (keypoint). Floating-point
descriptors are vectors of continuous numbers and are usually compared
with the $L_2$-norm. Binary descriptors are stored in binary
vectors and are compared using the Hamming distance $L_\mathcal{H}$~\cite{1950-hamming}.
Descriptors are computed so that the distance between them
grows with the dissimilarity between the corresponding, described image regions.

Binary descriptors generally occupy significantly less memory (128 to 512 bits)
than their floating-point counterparts (512 to 4096 bits) and are often cheaper to compute.
Furthermore, binary descriptors can be compared much faster than floating-point descriptors,
since the Hamming distance can be efficiently computed on modern CPUs.
Additionally, state-of-the-art binary descriptors
such as~\cite{2010-calonder-brief, 2011-rublee-orb, 2011-leutenegger-brisk, 2011-alcantarilla-fast, 2012-alahi-freak}
are more accessible than state-of-the-art floating point descriptors, which are subject to patents pending~\cite{2004-lowe-sift}.
For these reasons binary descriptors are generally preferred to floating-point descriptors
for \gls{vpr} or \gls{slam} applications~\cite{2017-mur-orbslam, 2017-pire-sptam, 2018-schlegel-proslam}.

Binary descriptors are computed based on local intensity
properties of the image, and are targeted at \emph{image recognition}.
They do not encode information originating from additional cues,
that are not necessarily present in the image.
In \gls{vpr} applications one often obtains such cues (e.g. point depth),
that can be used to verify recognized image candidates in a postprocessing phase.
In the following, we define an approach for adding such continuous or integer cues to binary descriptors,
to capture this additional information directly in the descriptor. 
The benefit of extending existing descriptors is that the remaining part of the system originally
thought to deal with binary descriptors does not have to be changed to work with the additional cues,
except for the length of the descriptors considered.

\subsection{Converting Continuous Cues into Binary Strings}\label{sec:converting-continuous-cues}

Without loss of generality, let $c$ be a continuous value in the
interval $\left[0, 1\right)$ encoding a cue that we want to add to a
binary descriptor $\descriptor$.  If $c$ is not contained in the
target range, one can use an affine operator
$\overline c = \alpha c + \beta$
such that the values of $\overline c$ lie in the specified
range.  For the sake of notation, in the remainder of this section
we assume $c$ to be normalized for the target range.

We aim at converting the value $c$ into a binary string
$\bb=b(c)$, that appended to the original descriptor $\descriptor$,
results in a new descriptor $\descriptor_\star=\left<\descriptor,
\bb\right>$.  Equally to $\descriptor$, also $\descriptor_\star$ is
compared with the Hamming distance.  Thus, $\bb$
must be Hamming distance compatible as well.  Furthermore, to transfer
the gathered discrimination between two values $c$ and $c'$, we need
to ensure that the Hamming distance between the binary strings $\bb$
and $\bb'$ computed from $c$ and $c'$, grows monotonically with the
distance between $c$ and $c'$.  These requirements can be
expressed by the following equation:
\begin{equation}
  L_\mathcal{H}(b(c), b(c')) \propto \left|c-c'\right|  \label{eqn:condition-continuous-embedding}.
\end{equation}

\begin{figure}[h!]
  \vspace{5pt}
  \centering
  \includegraphics[width=\columnwidth]{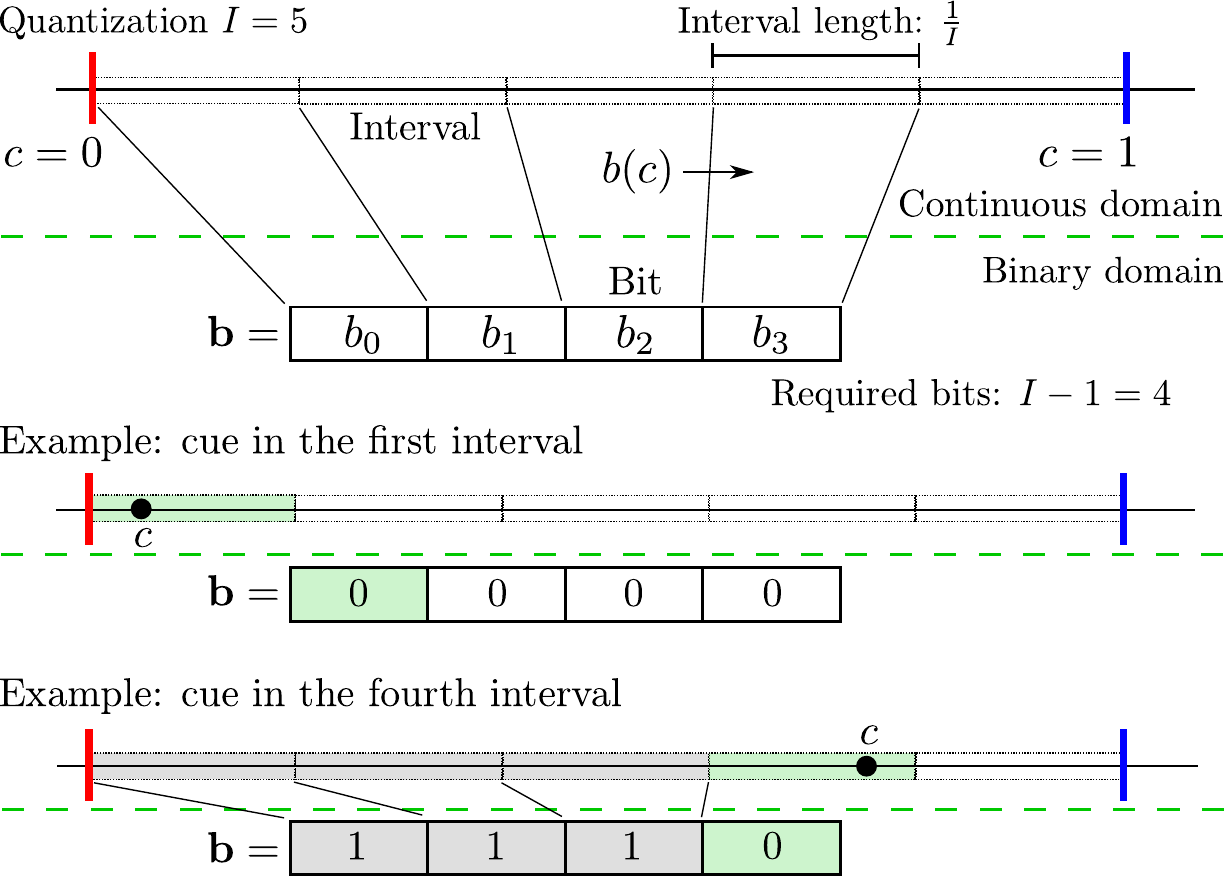}
  \vspace{-5pt}
  \caption{Top: Visualization of the proposed quantization and assignment function
  described in \eqref{eqn:continuous-embedding-function} with a quantization in $I=5$ intervals.
  Center and bottom: Example conversion for two different cue values.
  Note that we do not have to set any bit for cues in the first interval (center).}
  \label{fig:continuous-cue-n-dimension}
  \vspace{-0pt}
\end{figure}

To compute $b(c)$ we quantize the range $\left[0,1\right)$ in $I$ even intervals, each of length $1/I$.
For such a quantization, we can retrieve a binary string $\bb=b(c)$
consisting of $I-1$ bits, according to:
\begin{equation}
  b(c)=\left<b_0(c), b_1(c), \dots, b_{I-2}(c)\right>. \label{eqn:continuous-embedding-function}
\end{equation}
Where the value of the bit $b_i(c)$ is $1$ if $c$ lies in an interval higher than the interval at $i+1$, $0$ otherwise.
More formally:
\begin{equation}
  b_{i}(c)=
  \begin{cases} 
    1 & \text{iff}~c > \frac{1}{I} (i+1)\\
    0 & \text{otherwise}
  \end{cases}~i \in \left\{0, 1, \dots, I-2\right\}.
  \label{eqn:continuous-embedding-bit}
\end{equation}
\figref{fig:continuous-cue-n-dimension} illustrates the proposed procedure
with a quantization of $I=5$ intervals and two example cue conversions.

The above procedure can be applied also when the domain of the cues is
multi-dimensional. In that case, each cue $c_n$ will \emph{contribute with an
independent binary string} $\bb_n$ to $\descriptor_\star$.  The Hamming
distance between the augmented descriptors $\descriptor_\star,
\descriptor_\star'$ of multi-dimensional cues is proportional to
the Manhattan distance in the continuous space.
In the following, we provide a straightforward example for transforming two-dimensional cues
according to our procedure.

\emph{Example - Converting \gls{kc}:}
In relevant \gls{vpr} applications, such as autonomous cars,
images of the same place are often acquired from viewpoints similar to each other.
Feature detectors, such as the popular FAST corner detector~\cite{2006-rosten-fast-detector},
are constructed to return similar keypoint detections for similar images.
Hence, we analyze the effect of adding the keypoint coordinates $(u,v)$ to the descriptors w.r.t. the achieved \gls{vpr} accuracy.

\begin{figure}[h!]
  \centering
  \vspace{-0pt}
  \includegraphics[width=\columnwidth]{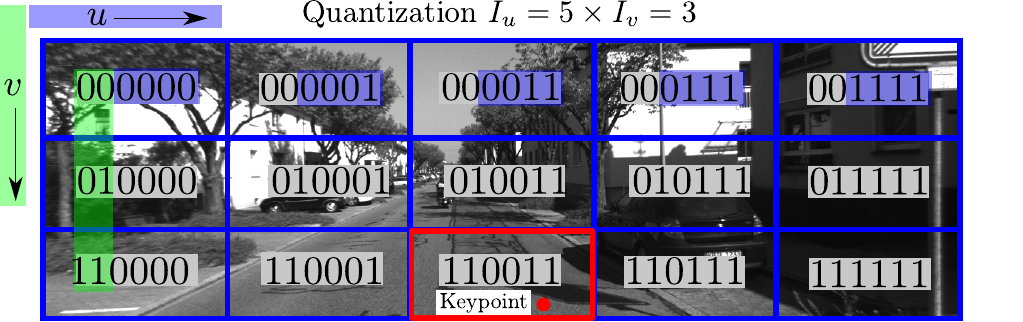}
  \vspace{-10pt}
  \caption{Proposed two-dimensional cue conversion based on a keypoint position $c=(u,v)$ (red dot)
  in a $I_u=3\times I_v=5$ quantization scenario.
  The obtained binary strings are $\bb_v=b(c_v)=\left<1,1\right>$ and $\bb_u=b(c_u)=\left<0,0,1,1\right>$,
  resulting in the composite string $\bb=\left<1,1,0,0,1,1\right>$ (red box).}
  \label{fig:position-augmentation}
  \vspace{-5pt}
\end{figure}

To convert this two-dimensional cue, we need only to define the vertical
quantization $I_v$ and horizontal quantization $I_u$ of the keypoint coordinates $(u,v)$.
In \figref{fig:position-augmentation} we display a two-dimensional
image quantization of $I_u=5, I_v=3$ and the resulting binary string
$\bb=b(c)$ computed with \eqref{eqn:continuous-embedding-function}, for
an example keypoint.

Depending on the chosen quantization, higher distances will result in the dimension with the most partitions.
This enables us to stress horizontal position similarities over vertical ones.
Notably, the quantization and the binary string mapping
can be computed once at startup for all possible cues (i.e. pixels) on the image plane.
This enables us to transform a keypoint position with a \emph{single lookup} at runtime.
We validate the effectiveness of our additional cue for \gls{vpr}
in a series of experiments (\secref{sec:experimental-evaluation}).

\subsection{Converting Selector Cues into Binary Strings}\label{sec:converting-selected-cues}

In the previous section we focused on embedding \emph{continuous values} into binary descriptors.
Here we describe a procedure to embed an arbitrary \emph{selector value}.

A selector value $i$ can be mapped to the finite integer range $\mathcal{I}=\left\{0, 1, \dots, I-1\right\}$
and typically encodes \emph{discrete cues} such as label information.
Let the distance between two selectors be either $2$ if they are different, or $0$ if they are identical.
This distance property can be straightforwardly implemented with the Hamming distance
by using a binary string $\bb$ of length $I$ bits.
A cue $c$ of value $i\in\mathcal{I}$ is represented by a binary string $\bb=\left<b_0(c), b_1(c), \dots, b_{I-1}(c)\right>$,
where the bit $b_i(c)$ is set to $1$, and all others to $0$.

\begin{figure}[h!]
  \centering
  \vspace{-0pt}
  \includegraphics[width=\columnwidth]{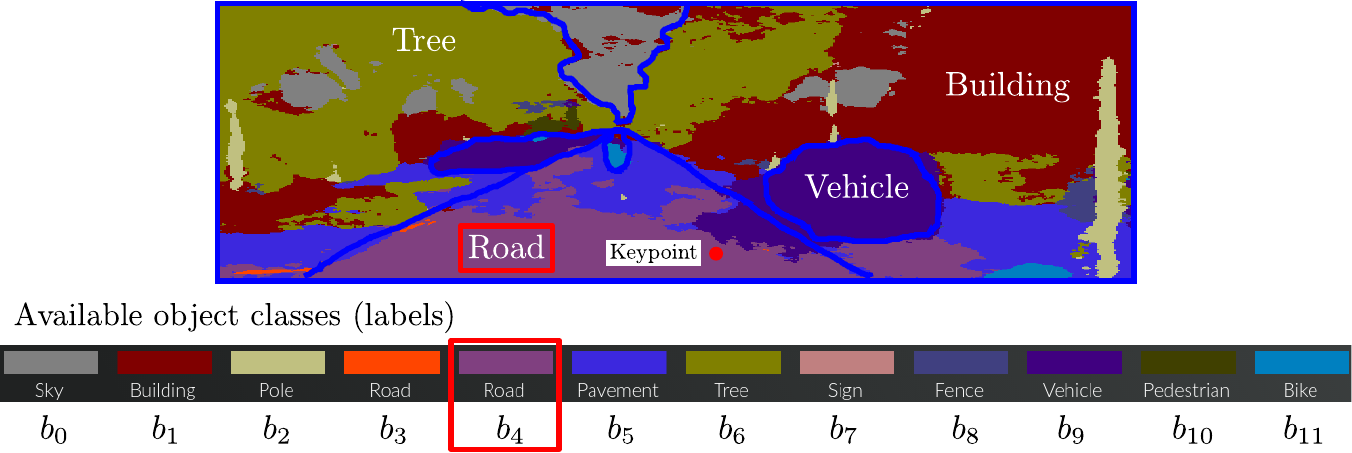}
  \vspace{-10pt}
  \caption{Proposed cue conversion based on the label ($1$ of $12$) at the keypoint location (red dot).
  In this scenario, the example keypoint is labeled as: Road,
  resulting in the string $\bb=\left<0,0,0,0,1,0,0,0,0,0,0,0\right>$ (red box).}
  \label{fig:augmentation-semantics}
  \vspace{-5pt}
\end{figure}

\emph{Example - Converting \gls{sl}:}
In the remainder of this section we provide an example on how to use selector cues
within a \gls{vpr} system based on binary descriptors.
To this extent we add to each descriptor the semantic label computed at its keypoint position.
We retrieve the label by using the SegNet~\cite{2015-badrinarayanan-segnet-cnn} system\footnote{SegNet: \url{www.github.com/alexgkendall/caffe-segnet}}.
We ran SegNet with an off-the-shelf configuration ($\mathrm{webdemo}$)
for which we can recognize up to $12$ selector cues.
\figref{fig:augmentation-semantics} illustrates the described selector mapping for an example image.

Albeit the segmentation was not always correct, the precision of the \gls{vpr} system improved,
as reported in our experiments (\secref{sec:experimental-evaluation}).
We remark that for \gls{vpr}, we do not necessarily rely on \emph{correct} labels
as long as the labeling is \emph{consistent} (e.g. if the network mistakenly labels a vehicle as a tree in a place,
and as we return to that place, does the same mistake again,
we still have consistent similarity search information). 
Consistency is easier to achieve than correctness, since the latter implies the former.

In contrast to the continuous cues of the \acrfull{kc} of the previous example,
the \gls{sl} labels are \emph{rotation and translation insensitive}.
Hence, images acquired from different viewpoints can still be matched.

\subsection{Augmentation Weighting}
When the Hamming distance $L_\mathcal{H}(\descriptor, \descriptor')$ between two descriptors $\descriptor, \descriptor'$ is smaller than a certain threshold $\tau\geq0$,
$\descriptor, \descriptor'$ are assumed to originate from the same image feature (\emph{match}).
Typical binary descriptors have bit sizes of $256$ or $512$.
Depending on the desired continuous range quantization,
the resulting, auxiliary binary string $\bb$ can be much smaller (e.g. $6$ bits) than the descriptor $\descriptor$. 
Hence, the augmented descriptor $\descriptor_\star=\left<\descriptor, \bb\right>$ might change only very little in size. 
In such a case, the cues' contributions to $\descriptor_\star$ respectively $\descriptor_\star'$ vanish when the Hamming distance
is computed, more formally: $L_\mathcal{H}(\descriptor_\star, \descriptor_\star') \approx L_\mathcal{H}(\descriptor, \descriptor')$.

To tackle this issue,
we introduce a variable \emph{augmentation weight} $\augmentationweight \geq 0$
and define the \emph{weighted} Hamming distance as:
\begin{equation}
  L_\mathcal{H}^\lambda(\descriptor_\star, \descriptor_\star')=L_\mathcal{H}(\descriptor, \descriptor')
                                                      +\augmentationweight L_\mathcal{H}(\bb, \bb')
  \label{eqn:weighted-hamming-distance}
\end{equation}
It is easy to see that for $\augmentationweight \to 0$ the contribution of the added cue $\bb$ vanishes,
whereas for $\augmentationweight \to \infty$ the original descriptor $\descriptor$ is not considered anymore.
Instead of modifying the search method to compute the weighted Hamming distance,
\eqref{eqn:weighted-hamming-distance} can also be achieved by appending
an augmentation $\augmentationweight$ times to a descriptor (assuming $\lambda$ is an integer).
In our experiments we conducted such an evaluation, without touching the comparison function of the search methods.
We list the values of $\augmentationweight$ which
we found working best for our two example cues \gls{kc} and \gls{sl}.

\section{Experimental Evaluation}\label{sec:experimental-evaluation}

The focus of this work is to improve the precision of feature-based similarity search methods for \gls{vpr}
by adding cues (\secref{sec:converting-continuous-cues},~\secref{sec:converting-selected-cues}) to descriptors
without requiring major adjustments for the search methods.
In the following we introduce the descriptors (\secref{sec:descriptor-types}),
the search methods (\secref{sec:similarity-search-methods}), the performance metrics (\secref{sec:performance-metrics})
and the various datasets (\secref{sec:datasets}) considered in our experiments.
Subsequently, we display and discuss our results (\secref{sec:results}).

\subsection{Descriptors}\label{sec:descriptor-types}

Our cue embedding strategies (\secref{sec:converting-continuous-cues},~\secref{sec:converting-selected-cues})
and their beneficial effects for a similarity search method 
are independent of the selected binary descriptor.
To support this claim, we conducted experiments on several of the most common local binary descriptors:

\textbf{BRIEF}~\cite{2010-calonder-brief} was one of the first local binary descriptors
to enable efficient computation, storage and comparison.
Due to its viability for real-time applications, BRIEF is broadly used to this day.

\textbf{ORB}~\cite{2011-rublee-orb} improved BRIEF by adding rotational and scale invariance.
It is widely used in feature-based \gls{vo} systems.

\textbf{BRISK}~\cite{2011-leutenegger-brisk} improved BRIEF by considering a Gaussian pixel average for the descriptor computation.

\textbf{A-KAZE}~\cite{2011-alcantarilla-fast} is an accelerated version of the KAZE descriptor.
KAZE captures image regions in a nonlinear scale space, which comes at a high computational cost.
By using a more recent scheme to build the nonlinear scale space, the significant acceleration was obtained.

\textbf{FREAK}~\cite{2012-alahi-freak} is one of few bio-inspired descriptors.
Its binary signature is computed by efficiently comparing image intensities
over a retinal sampling pattern.

\textbf{LDAHash}~\cite{2012-strecha-ldahash} compresses SIFT~\cite{2004-lowe-sift} descriptors in binary descriptors
through a supervised binarization scheme.

\textbf{BinBoost}~\cite{2013-trzcinski-boosting} is an extremely compact binary descriptor (as small as 8 bits)
where each individual bit is computed by a learned hash function.
The hash functions are learned such that each bit complements the others.


We utilized the current OpenCV\footnote{OpenCV library: \url{www.opencv.org}} (3.4.7) implementation
of all descriptor types except for LDAHash where we integrated the released C++ source code by the authors.

\subsection{Similarity Search Methods}\label{sec:similarity-search-methods}

We tested our augmented descriptors on 4 state-of-the art feature-based similarity search methods
(\secref{sec:related-work}).
Each one of them having its advantages and drawbacks.

\textbf{\gls{bf}}:
The straightforward, exhaustive search approach, guaranteeing the best matches at the highest computational cost.
We utilized the official implementation of the publicly available OpenCV library (3.4.7).

\textbf{\gls{lsh}}:
Fast Library for Approximate Nearest Neighbors with multi-probe Locality-sensitive hashing.
We utilized the official implementation of the OpenCV library (3.4.7).

\textbf{\gls{bof}}:
State-of-the-art method for fast place recognition, employed by many current \gls{slam} systems.
A well maintained and widely used \gls{bof} library in conjunction with binary descriptors is \gls{dbow}~\cite{2012-galvez-bow}.
We utilized the authors' publicly available implementation\footnote{DBoW2 library: \url{www.github.com/dorian3d/DBoW2}}.

\textbf{\gls{bst}}:
A binary search tree approach, suitable for highly efficient
and lightweight similarity search in large scale datasets with binary descriptors.
We utilized the publicly available \gls{hbst}~\cite{2018-schlegel-hbst} implementation\footnote{HBST library: \url{www.gitlab.com/srrg-software/srrg_hbst}}.

\subsection{Performance Metrics}\label{sec:performance-metrics}

We chose to examine multiple of the most common performance metrics used in \gls{vpr}:

\textbf{\acrfull{pr}}:
To determine the reliability of a \gls{vpr} system
one generally measures the resulting Precision and Recall statistics.
The first being:
\begin{equation}
Precision = \frac{\#~correctly~reported~matches}{\#~total~reported~matches} \in \left[0,1\right]. \notag
\end{equation}
Regarding the completeness of the results, one considers:
\begin{equation}
Recall = \frac{\#~correctly~reported~matches}{\#~total~possible~correct~matches} \in \left[0,1\right]. \notag
\end{equation}
Here, \emph{match} refers to a pair of images that are reported by the \gls{vpr} system to originate from the same place.

\textbf{\gls{map}}: The \gls{map} reflects the average Precision of \emph{multiple} \gls{pr} curves over equidistant Recall levels.
With the \gls{map} one can concisely describe the achieved performance of an approach over a \emph{series of test cases}
in a single number.


\textbf{Mean image processing time $\overline{t}$}:
To examine the impact of augmented descriptors on the computational cost
of a search method, we measure the required time for matching a set of descriptors of an image and integrating it into the database.

For the Oxford and the Paris datasets (\secref{sec:datasets}),
we concisely present our results using \gls{map}
(for which the authors specifically provide a tool~\cite{2007-philbin-object}).
We adapted the tool to also compute the \gls{map} for the ZuBuD and the Holidays dataset.
For KITTI we present individual \gls{pr} curves.

\subsection{Datasets}\label{sec:datasets}

For evaluating feature-based \gls{vpr} similarity search methods,
one generally considers a large number of reference images,
of which only few describe the same place as a query image.
Many of the images serve purely to perturb the search method (\emph{distractor images}).
Ground truth information is generally provided by listing the matching query to reference image pairs (\emph{correctly reported matches} / \emph{true positives}).
This data is needed to compute the achieved \gls{pr} and \gls{map} performance indicators (\secref{sec:performance-metrics}).

\begin{figure}[h!]
  \centering
  \vspace{-0pt}
  \begin{subfigure}[t]{0.385\columnwidth}
    \includegraphics[height=70pt]{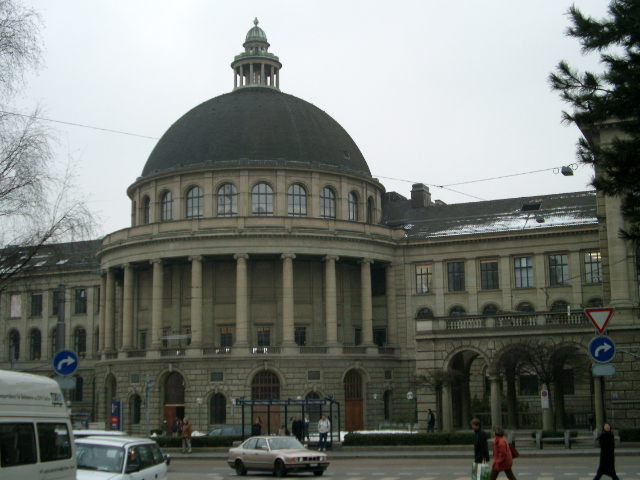}
    \subcaption{ZuBuD~\cite{2003-shao-zubud}}
  \end{subfigure}
  \begin{subfigure}[t]{0.215\columnwidth}
    \includegraphics[height=70pt]{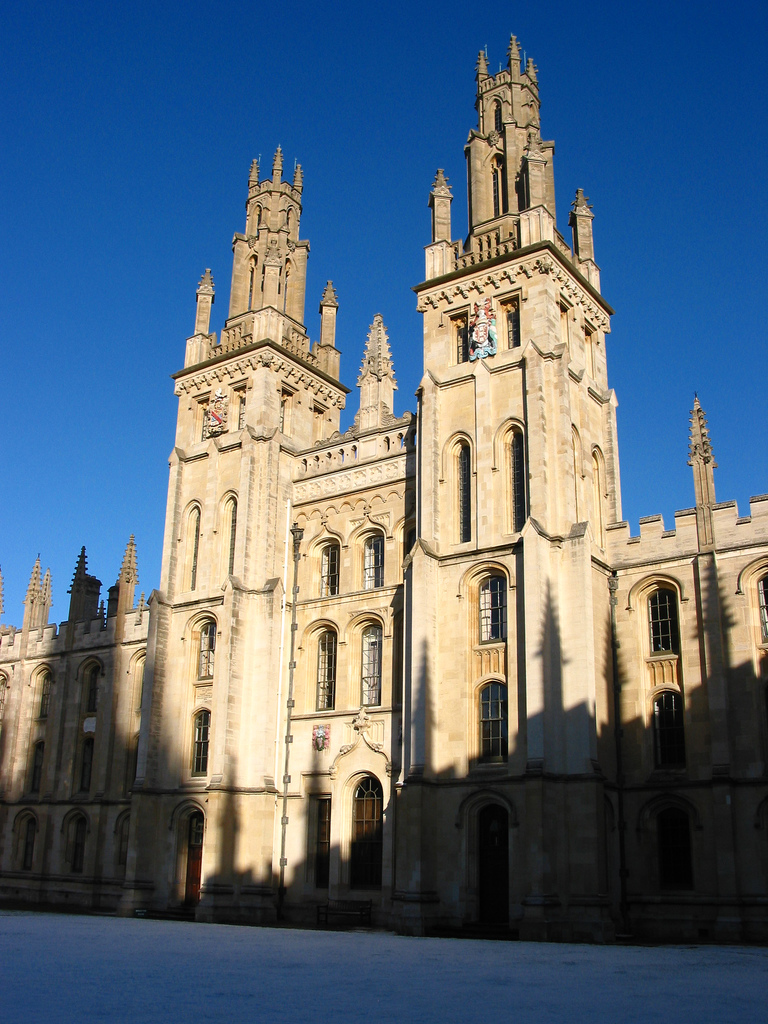}
    \subcaption{Oxford~\cite{2007-philbin-object}}
  \end{subfigure}
  \vspace{5pt}
  \begin{subfigure}[t]{0.37\columnwidth}
    \includegraphics[height=70pt]{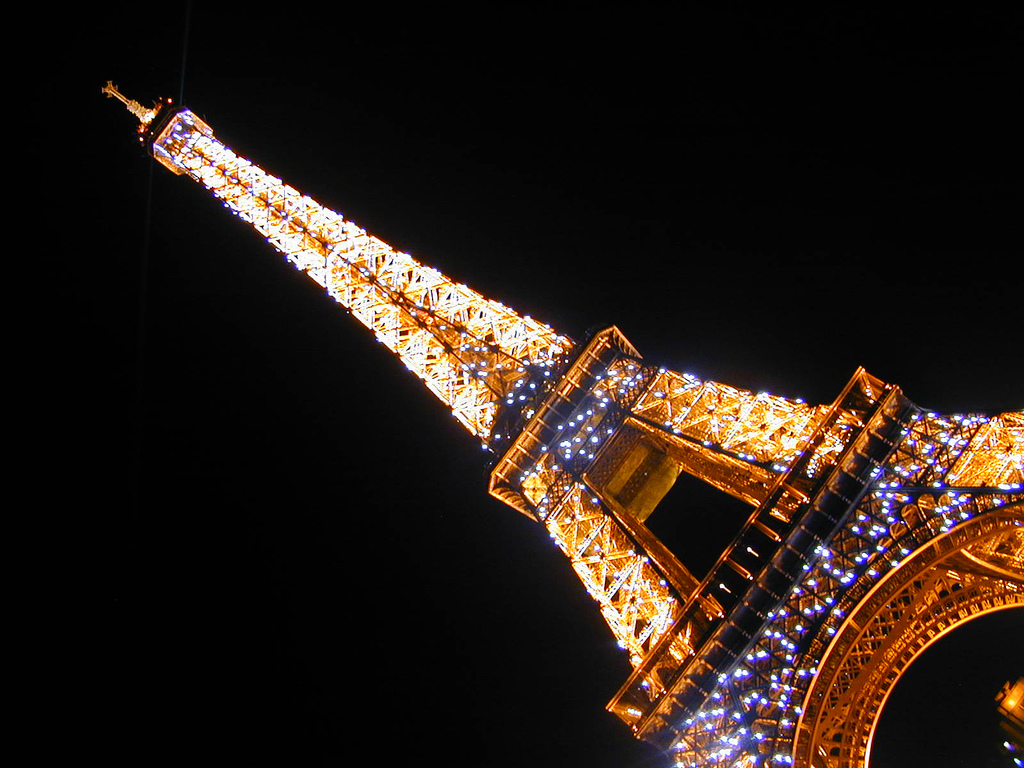}
    \subcaption{Paris~\cite{2008-philbin-lost}}
  \end{subfigure}
  \begin{subfigure}[b]{0.287\columnwidth}
    \includegraphics[height=52.5pt]{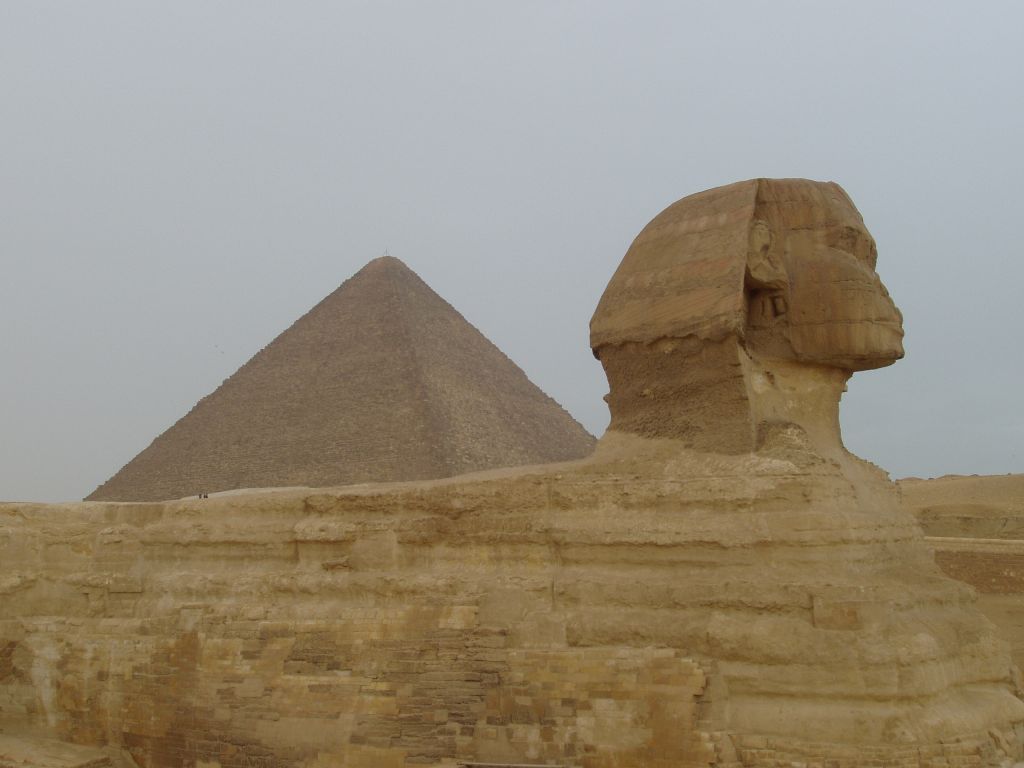}
    \subcaption{Holidays~\cite{2008-jegou-hamming}}
  \end{subfigure}
  \begin{subfigure}[b]{0.7\columnwidth}
    \includegraphics[height=52.5pt]{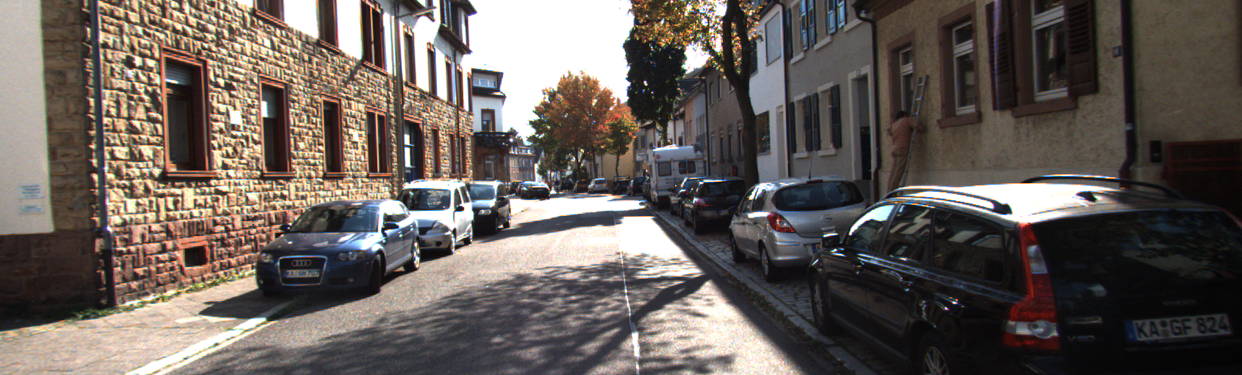}
    \subcaption{KITTI~\cite{2012-geiger-kitti}}
  \end{subfigure}
  \caption{Example images from our selected \gls{vpr} benchmark datasets.}
  \vspace{-5pt}
\label{fig:datasets}
\end{figure}

\figref{fig:datasets} displays an image sample for each of the five datasets we considered.
All of them are publicly approved and well-used standard datasets
for evaluating the performance of a \gls{vpr} system. 

Each one of the five datasets brings along a different set of particularities and challenges:

\paragraph{\textbf{ZuBuD (Zurich Buildings Database)}}
ZuBuD~\cite{2003-shao-zubud} contains $115$ query and $1005$ reference images of city buildings in Zurich.
The high similarity of the buildings challenges the discretization capabilities of search methods.

\paragraph{\textbf{Oxford Buildings Dataset}}
This dataset~\cite{2007-philbin-object} consists of $55$ query and $5008$ reference images collected from Flickr by searching for Oxford landmarks.
It has been manually annotated to generate a reliable ground truth. 

\paragraph{\textbf{Paris Dataset}}
The Paris Dataset~\cite{2008-philbin-lost} consists of $55$ query and $6357$ reference images collected from Flickr by searching for landmarks in Paris.
Many shots have been taken at night, significantly complicating the matching.

\paragraph{\textbf{Holidays Dataset}}
The Holidays dataset~\cite{2008-jegou-hamming} is a set of $500$ query and $991$ reference images, many captured with high camera resolutions ($3$-$6$ megapixels).
It includes a large variety of scene types (natural, man-made, etc.).

\paragraph{\textbf{KITTI \gls{vo} / \gls{slam} Evaluation 2012}}
The popular benchmark dataset~\cite{2012-geiger-kitti} for \gls{vo} and \gls{slam} approaches
contains 6 loop closure sequences with ground truth trajectories.
We computed an image matching ground truth using the provided trajectories and geometric verification.

\begin{figure*}[ht!]
  \centering
  \adjustbox{width=0.45\columnwidth, trim=59pt 0pt 60pt 10pt, clip}{
  \begin{picture}(7200.00,5040.00)%
      \gdef\gplbacktext{}
    \gdef\gplfronttext{}
    \gplgaddtomacro\gplbacktext{%
      \csname LTb\endcsname%
      \put(1905,949){\makebox(0,0)[r]{\strut{}$0.1$}}%
      \csname LTb\endcsname%
      \put(1905,1439){\makebox(0,0)[r]{\strut{}$0.2$}}%
      \csname LTb\endcsname%
      \put(1905,1929){\makebox(0,0)[r]{\strut{}$0.3$}}%
      \csname LTb\endcsname%
      \put(1905,2419){\makebox(0,0)[r]{\strut{}$0.4$}}%
      \csname LTb\endcsname%
      \put(1905,2909){\makebox(0,0)[r]{\strut{}$0.5$}}%
      \csname LTb\endcsname%
      \put(1905,3399){\makebox(0,0)[r]{\strut{}$0.6$}}%
      \csname LTb\endcsname%
      \put(1905,3889){\makebox(0,0)[r]{\strut{}$0.7$}}%
      \csname LTb\endcsname%
      \put(1905,4379){\makebox(0,0)[r]{\strut{}$0.8$}}%
      \csname LTb\endcsname%
      \put(2037,484){\makebox(0,0){\strut{}0}}%
      \csname LTb\endcsname%
      \put(2650,484){\makebox(0,0){\strut{}1}}%
      \csname LTb\endcsname%
      \put(3262,484){\makebox(0,0){\strut{}2}}%
      \csname LTb\endcsname%
      \put(3875,484){\makebox(0,0){\strut{}4}}%
      \csname LTb\endcsname%
      \put(4487,484){\makebox(0,0){\strut{}8}}%
      \csname LTb\endcsname%
      \put(5100,484){\makebox(0,0){\strut{}16}}%
      \csname LTb\endcsname%
      \put(5712,484){\makebox(0,0){\strut{}32}}%
    }%
    \gplgaddtomacro\gplfronttext{%
      \csname LTb\endcsname%
      \put(1267,2541){\rotatebox{-270}{\makebox(0,0){\strut{}mAP}}}%
      \put(3874,154){\makebox(0,0){\strut{}Augmentation weight $\lambda$}}%
      \put(3874,4709){\makebox(0,0){\strut{}Precision ZuBuD: BF with KC-8x8}}%
    }%
    \gplbacktext
    \put(0,0){\includegraphics{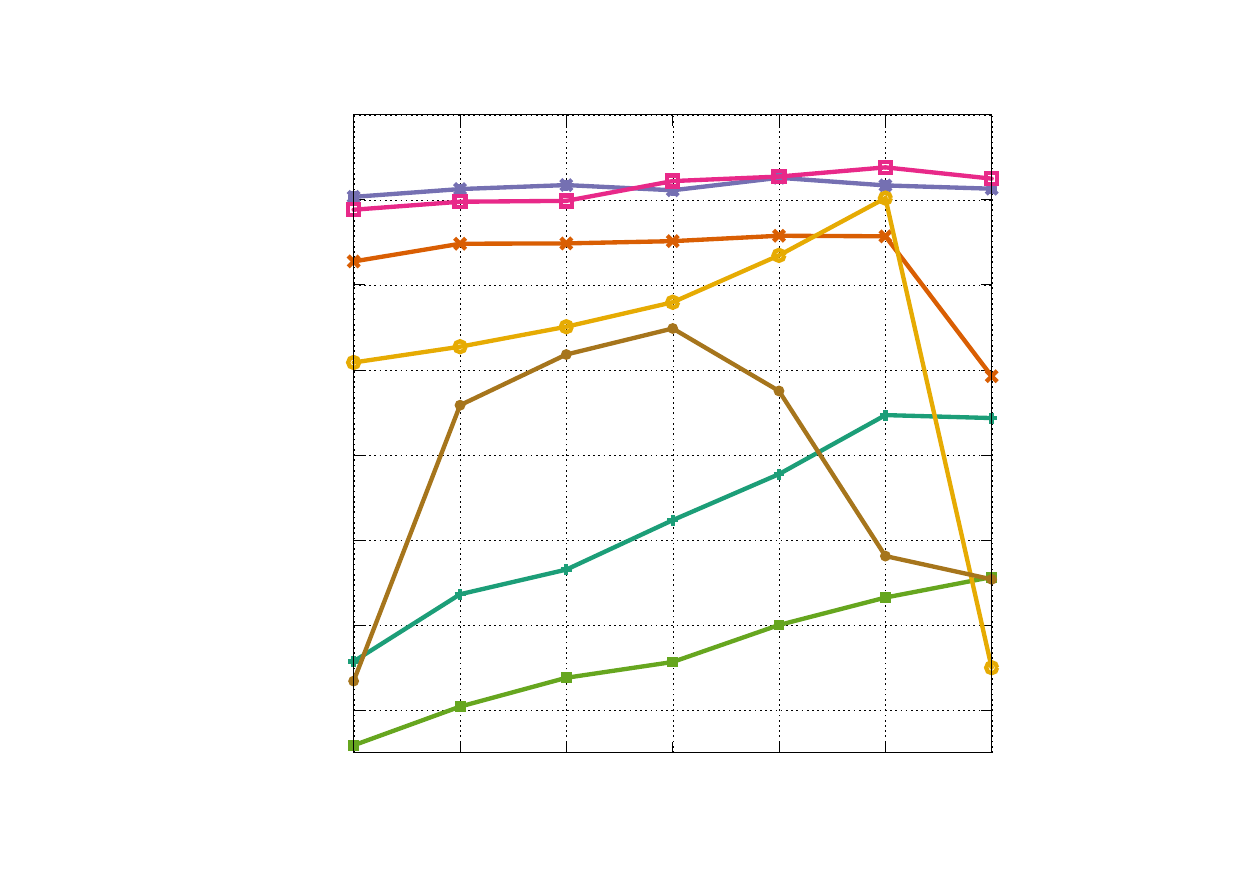}}%
    \gplfronttext
  \end{picture}
  }
  \adjustbox{width=0.45\columnwidth, trim=59pt 0pt 60pt 10pt, clip}{
  \begin{picture}(7200.00,5040.00)%
      \gdef\gplbacktext{}
    \gdef\gplfronttext{}
    \gplgaddtomacro\gplbacktext{%
      \csname LTb\endcsname%
      \put(1905,704){\makebox(0,0)[r]{\strut{}$0$}}%
      \csname LTb\endcsname%
      \put(1905,1754){\makebox(0,0)[r]{\strut{}$0.1$}}%
      \csname LTb\endcsname%
      \put(1905,2804){\makebox(0,0)[r]{\strut{}$0.2$}}%
      \csname LTb\endcsname%
      \put(1905,3854){\makebox(0,0)[r]{\strut{}$0.3$}}%
      \csname LTb\endcsname%
      \put(2037,484){\makebox(0,0){\strut{}0}}%
      \csname LTb\endcsname%
      \put(2650,484){\makebox(0,0){\strut{}1}}%
      \csname LTb\endcsname%
      \put(3262,484){\makebox(0,0){\strut{}2}}%
      \csname LTb\endcsname%
      \put(3875,484){\makebox(0,0){\strut{}4}}%
      \csname LTb\endcsname%
      \put(4487,484){\makebox(0,0){\strut{}8}}%
      \csname LTb\endcsname%
      \put(5100,484){\makebox(0,0){\strut{}16}}%
      \csname LTb\endcsname%
      \put(5712,484){\makebox(0,0){\strut{}32}}%
    }%
    \gplgaddtomacro\gplfronttext{%
      \csname LTb\endcsname%
      \put(1267,2541){\rotatebox{-270}{\makebox(0,0){\strut{}mAP}}}%
      \put(3874,154){\makebox(0,0){\strut{}Augmentation weight $\lambda$}}%
      \put(3874,4709){\makebox(0,0){\strut{}Precision Oxford: LSH with KC-8x8}}%
    }%
    \gplbacktext
    \put(0,0){\includegraphics{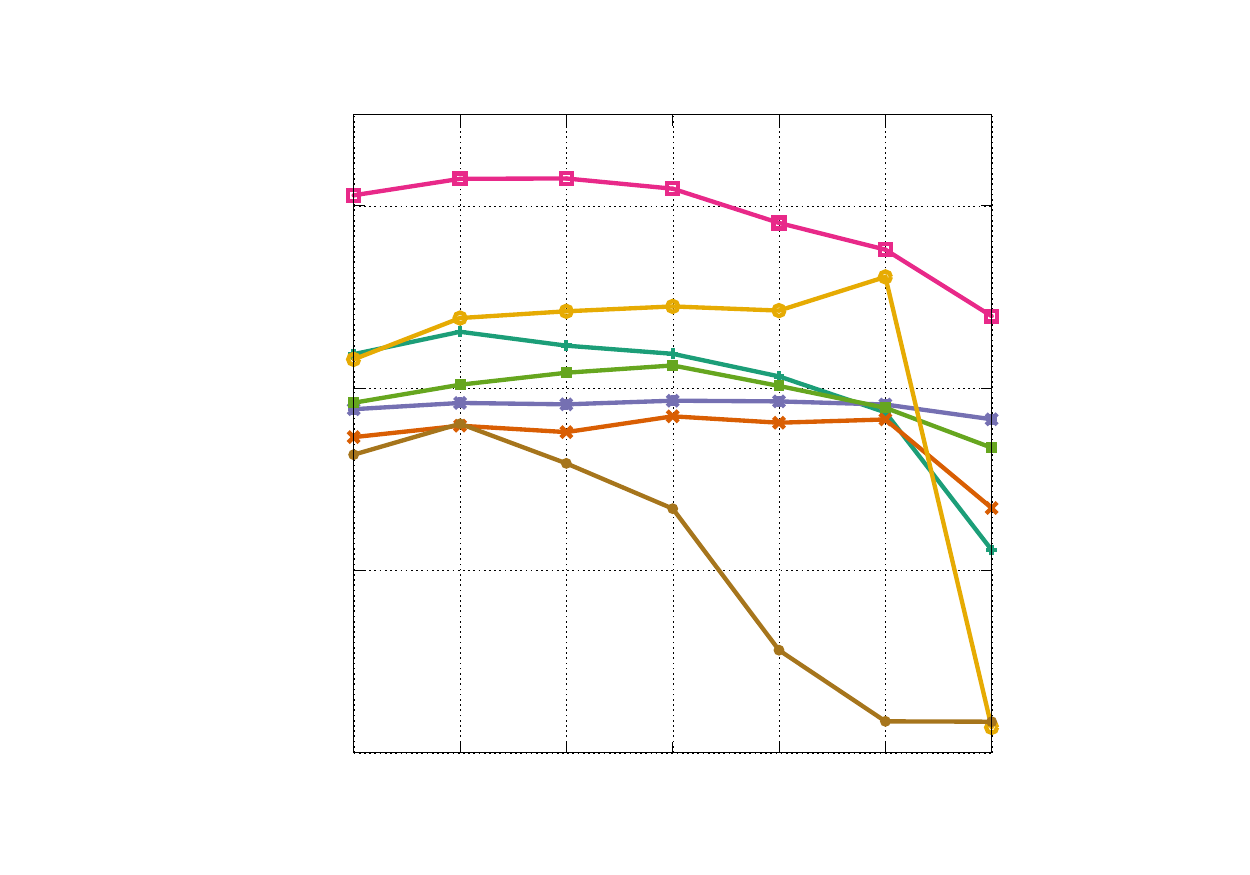}}%
    \gplfronttext
  \end{picture}
  }
  \adjustbox{width=0.45\columnwidth, trim=59pt 0pt 60pt 10pt, clip}{
  \begin{picture}(7200.00,5040.00)%
      \gdef\gplbacktext{}
    \gdef\gplfronttext{}
    \gplgaddtomacro\gplbacktext{%
      \csname LTb\endcsname%
      \put(1905,1317){\makebox(0,0)[r]{\strut{}$0.1$}}%
      \csname LTb\endcsname%
      \put(1905,3766){\makebox(0,0)[r]{\strut{}$0.2$}}%
      \csname LTb\endcsname%
      \put(2037,484){\makebox(0,0){\strut{}0}}%
      \csname LTb\endcsname%
      \put(2650,484){\makebox(0,0){\strut{}1}}%
      \csname LTb\endcsname%
      \put(3262,484){\makebox(0,0){\strut{}2}}%
      \csname LTb\endcsname%
      \put(3875,484){\makebox(0,0){\strut{}4}}%
      \csname LTb\endcsname%
      \put(4487,484){\makebox(0,0){\strut{}8}}%
      \csname LTb\endcsname%
      \put(5100,484){\makebox(0,0){\strut{}16}}%
      \csname LTb\endcsname%
      \put(5712,484){\makebox(0,0){\strut{}32}}%
    }%
    \gplgaddtomacro\gplfronttext{%
      \csname LTb\endcsname%
      \put(1267,2541){\rotatebox{-270}{\makebox(0,0){\strut{}mAP}}}%
      \put(3874,154){\makebox(0,0){\strut{}Augmentation weight $\lambda$}}%
      \put(3874,4709){\makebox(0,0){\strut{}Precision Paris: BOF with KC-8x8}}%
    }%
    \gplbacktext
    \put(0,0){\includegraphics{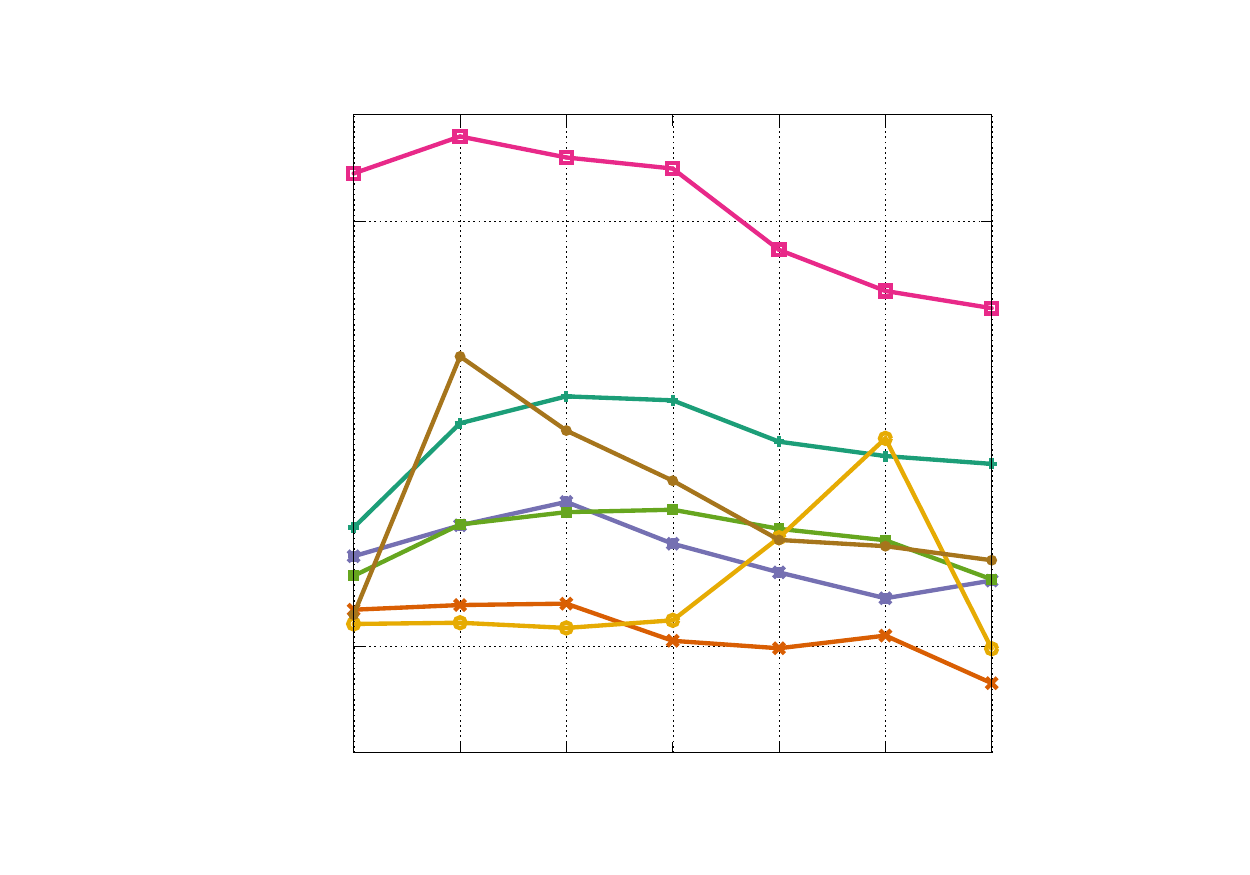}}%
    \gplfronttext
  \end{picture}
  }
  \adjustbox{width=0.45\columnwidth, trim=59pt 0pt 60pt 10pt, clip}{
  \begin{picture}(7200.00,5040.00)%
      \gdef\gplbacktext{}
    \gdef\gplfronttext{}
    \gplgaddtomacro\gplbacktext{%
      \csname LTb\endcsname%
      \put(1905,1112){\makebox(0,0)[r]{\strut{}$0.2$}}%
      \csname LTb\endcsname%
      \put(1905,1929){\makebox(0,0)[r]{\strut{}$0.3$}}%
      \csname LTb\endcsname%
      \put(1905,2746){\makebox(0,0)[r]{\strut{}$0.4$}}%
      \csname LTb\endcsname%
      \put(1905,3562){\makebox(0,0)[r]{\strut{}$0.5$}}%
      \csname LTb\endcsname%
      \put(1905,4379){\makebox(0,0)[r]{\strut{}$0.6$}}%
      \csname LTb\endcsname%
      \put(2037,484){\makebox(0,0){\strut{}0}}%
      \csname LTb\endcsname%
      \put(2650,484){\makebox(0,0){\strut{}1}}%
      \csname LTb\endcsname%
      \put(3262,484){\makebox(0,0){\strut{}2}}%
      \csname LTb\endcsname%
      \put(3875,484){\makebox(0,0){\strut{}4}}%
      \csname LTb\endcsname%
      \put(4487,484){\makebox(0,0){\strut{}8}}%
      \csname LTb\endcsname%
      \put(5100,484){\makebox(0,0){\strut{}16}}%
      \csname LTb\endcsname%
      \put(5712,484){\makebox(0,0){\strut{}32}}%
    }%
    \gplgaddtomacro\gplfronttext{%
      \csname LTb\endcsname%
      \put(1267,2541){\rotatebox{-270}{\makebox(0,0){\strut{}mAP}}}%
      \put(3874,154){\makebox(0,0){\strut{}Augmentation weight $\lambda$}}%
      \put(3874,4709){\makebox(0,0){\strut{}Precision Holidays: BST with KC-8x8}}%
    }%
    \gplbacktext
    \put(0,0){\includegraphics{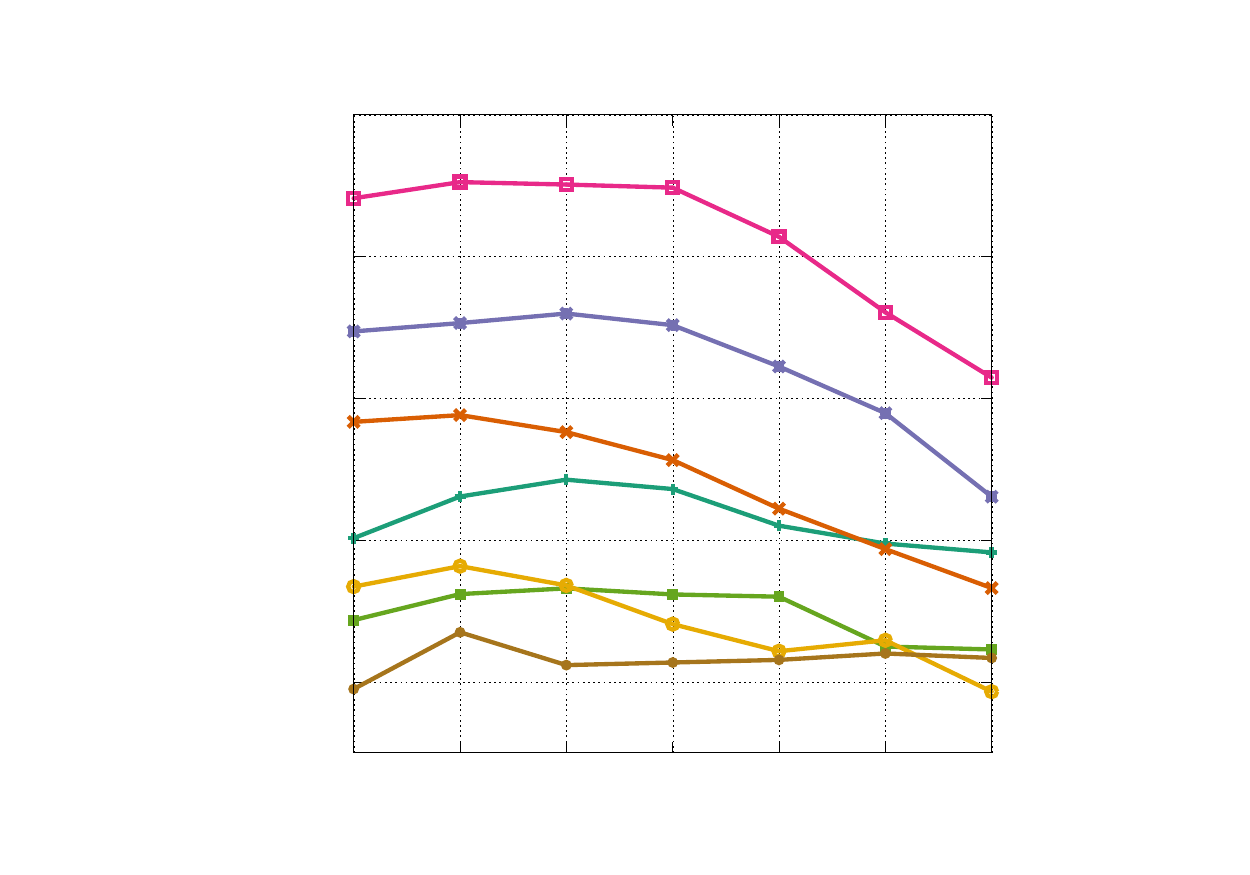}}%
    \gplfronttext
  \end{picture}
  }
  \adjustbox{width=0.2\columnwidth, trim=10pt 60pt 265pt 45pt, clip}{
    \begin{picture}(7200.00,5040.00)%
    \gdef\gplbacktext{}
    \gdef\gplfronttext{}
    \gplgaddtomacro\gplbacktext{%
    }%
    \gplgaddtomacro\gplfronttext{%
      \csname LTb\endcsname%
      \put(723,4011){\makebox(0,0)[l]{\strut{}BRIEF-256}}%
      \csname LTb\endcsname%
      \put(723,3791){\makebox(0,0)[l]{\strut{}ORB-256}}%
      \csname LTb\endcsname%
      \put(723,3571){\makebox(0,0)[l]{\strut{}BRISK-512}}%
      \csname LTb\endcsname%
      \put(723,3351){\makebox(0,0)[l]{\strut{}A-KAZE-486}}%
      \csname LTb\endcsname%
      \put(723,3131){\makebox(0,0)[l]{\strut{}FREAK-512}}%
      \csname LTb\endcsname%
      \put(723,2911){\makebox(0,0)[l]{\strut{}LDAHash-128}}%
      \csname LTb\endcsname%
      \put(723,2691){\makebox(0,0)[l]{\strut{}BinBoost-064}}%
    }%
    \gplbacktext
    \put(0,0){\includegraphics{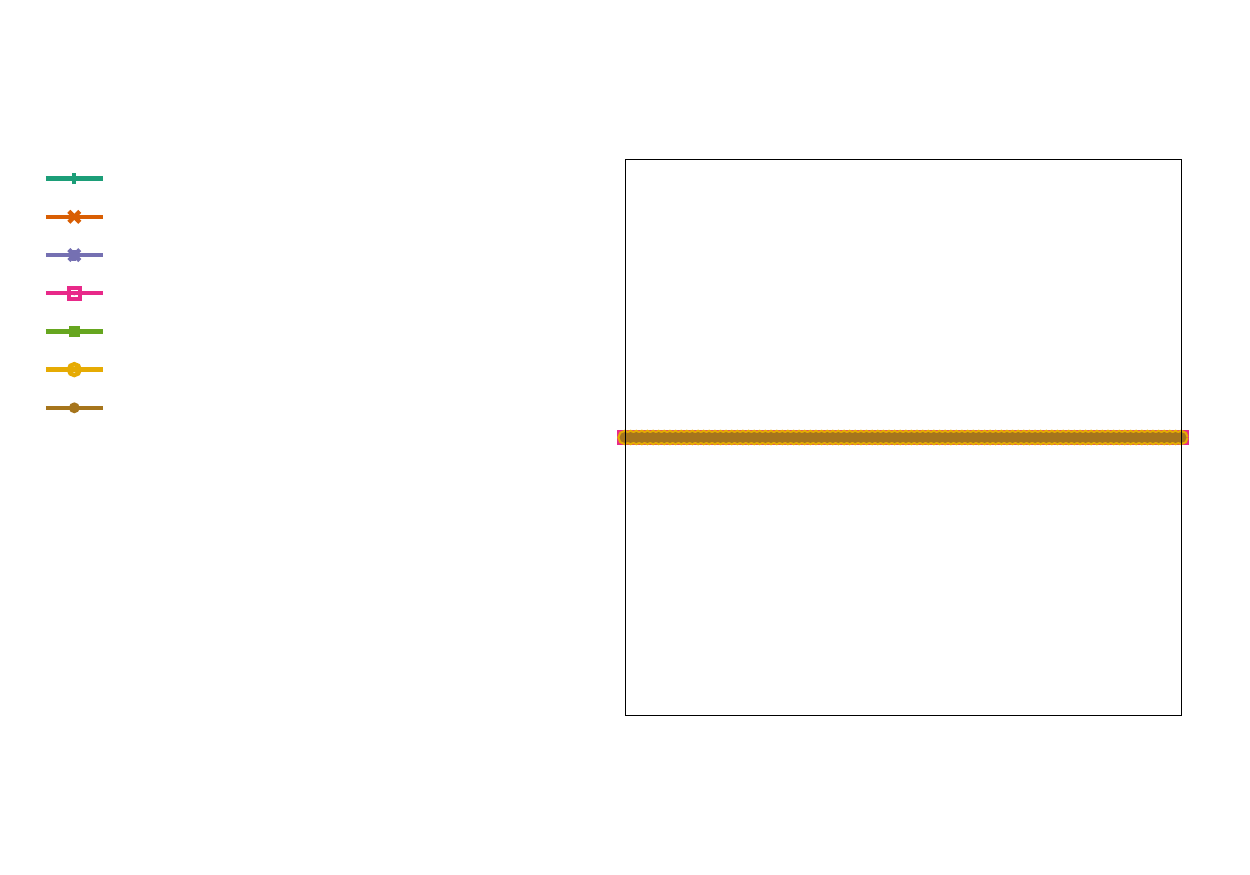}}%
    \gplfronttext
  \end{picture} 
  }
  \caption{Results for \gls{kc}-8x8 with varying augmentation weight $\lambda$. A maximum of $1000$ descriptors has been computed for each image.}
  \label{fig:results-kc}
  \vspace{-5pt}
\end{figure*}
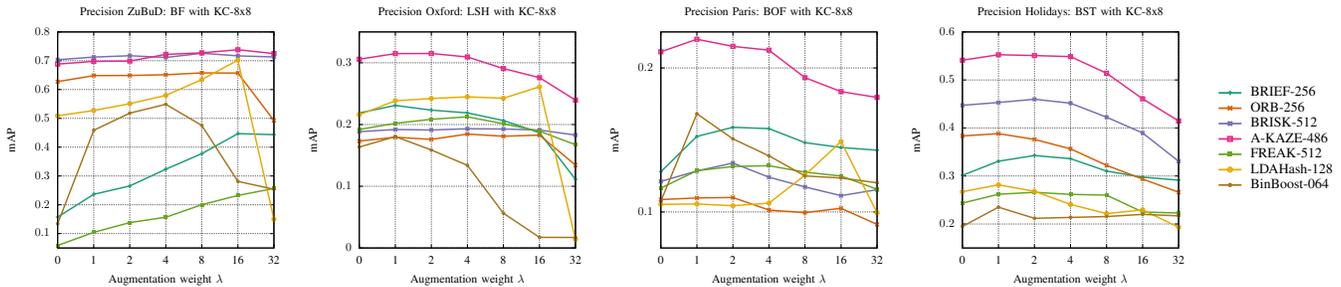
\begin{figure*}[ht!]
  \centering
  \adjustbox{width=0.45\columnwidth, trim=59pt 0pt 60pt 10pt, clip}{
  \begin{picture}(7200.00,5040.00)%
      \gdef\gplbacktext{}
    \gdef\gplfronttext{}
    \gplgaddtomacro\gplbacktext{%
      \csname LTb\endcsname%
      \put(1905,934){\makebox(0,0)[r]{\strut{}$0.1$}}%
      \csname LTb\endcsname%
      \put(1905,1393){\makebox(0,0)[r]{\strut{}$0.2$}}%
      \csname LTb\endcsname%
      \put(1905,1852){\makebox(0,0)[r]{\strut{}$0.3$}}%
      \csname LTb\endcsname%
      \put(1905,2312){\makebox(0,0)[r]{\strut{}$0.4$}}%
      \csname LTb\endcsname%
      \put(1905,2771){\makebox(0,0)[r]{\strut{}$0.5$}}%
      \csname LTb\endcsname%
      \put(1905,3231){\makebox(0,0)[r]{\strut{}$0.6$}}%
      \csname LTb\endcsname%
      \put(1905,3690){\makebox(0,0)[r]{\strut{}$0.7$}}%
      \csname LTb\endcsname%
      \put(1905,4149){\makebox(0,0)[r]{\strut{}$0.8$}}%
      \csname LTb\endcsname%
      \put(2037,484){\makebox(0,0){\strut{}0}}%
      \csname LTb\endcsname%
      \put(2650,484){\makebox(0,0){\strut{}4}}%
      \csname LTb\endcsname%
      \put(3262,484){\makebox(0,0){\strut{}8}}%
      \csname LTb\endcsname%
      \put(3875,484){\makebox(0,0){\strut{}16}}%
      \csname LTb\endcsname%
      \put(4487,484){\makebox(0,0){\strut{}32}}%
      \csname LTb\endcsname%
      \put(5100,484){\makebox(0,0){\strut{}64}}%
      \csname LTb\endcsname%
      \put(5712,484){\makebox(0,0){\strut{}128}}%
    }%
    \gplgaddtomacro\gplfronttext{%
      \csname LTb\endcsname%
      \put(1267,2541){\rotatebox{-270}{\makebox(0,0){\strut{}mAP}}}%
      \put(3874,154){\makebox(0,0){\strut{}Augmentation weight $\lambda$}}%
      \put(3874,4709){\makebox(0,0){\strut{}Precision ZuBuD: BF with SL-12}}%
    }%
    \gplbacktext
    \put(0,0){\includegraphics{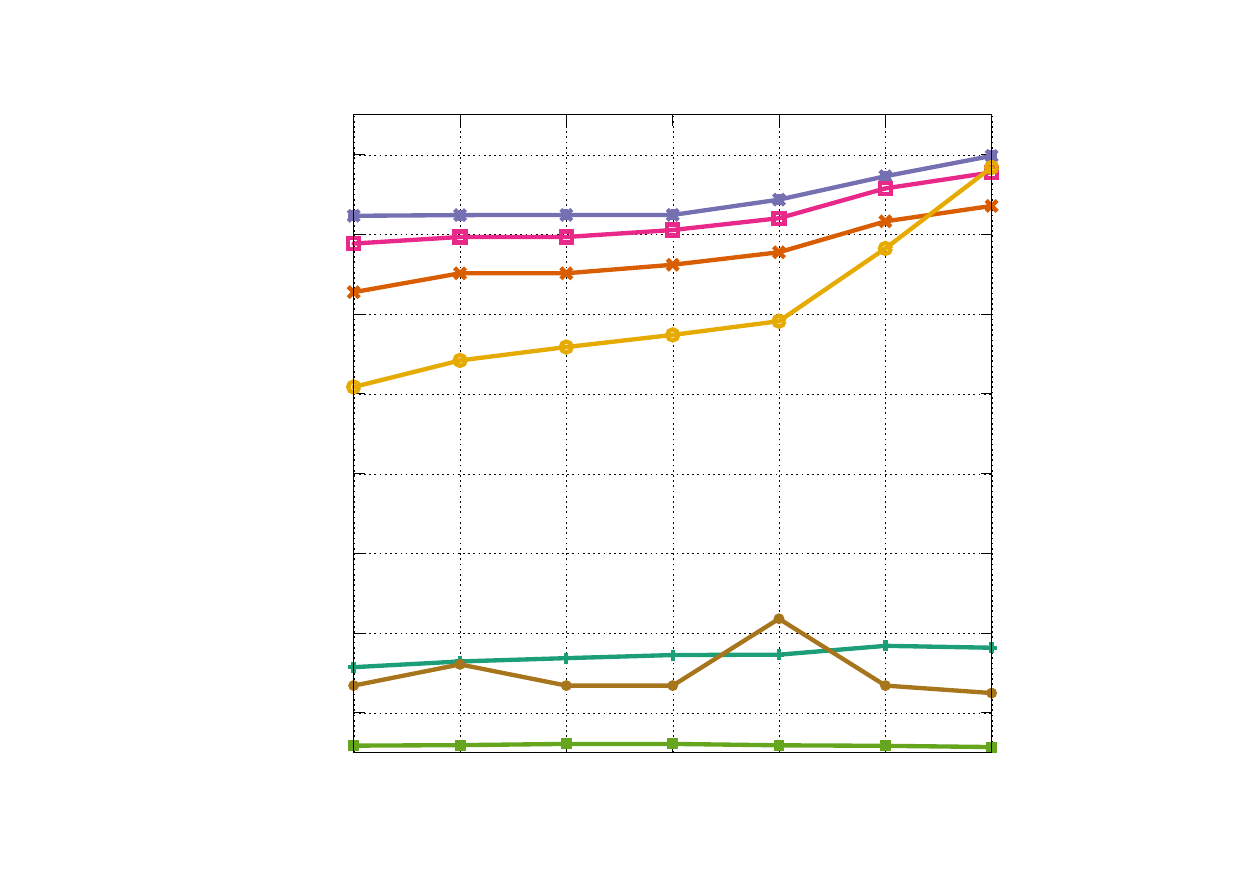}}%
    \gplfronttext
  \end{picture}
  }
  \adjustbox{width=0.45\columnwidth, trim=59pt 0pt 60pt 10pt, clip}{
  \begin{picture}(7200.00,5040.00)%
      \gdef\gplbacktext{}
    \gdef\gplfronttext{}
    \gplgaddtomacro\gplbacktext{%
      \csname LTb\endcsname%
      \put(1905,704){\makebox(0,0)[r]{\strut{}$0.1$}}%
      \csname LTb\endcsname%
      \put(1905,2542){\makebox(0,0)[r]{\strut{}$0.2$}}%
      \csname LTb\endcsname%
      \put(1905,4379){\makebox(0,0)[r]{\strut{}$0.3$}}%
      \csname LTb\endcsname%
      \put(2037,484){\makebox(0,0){\strut{}0}}%
      \csname LTb\endcsname%
      \put(3262,484){\makebox(0,0){\strut{}1}}%
      \csname LTb\endcsname%
      \put(4487,484){\makebox(0,0){\strut{}2}}%
      \csname LTb\endcsname%
      \put(5712,484){\makebox(0,0){\strut{}4}}%
    }%
    \gplgaddtomacro\gplfronttext{%
      \csname LTb\endcsname%
      \put(1267,2541){\rotatebox{-270}{\makebox(0,0){\strut{}mAP}}}%
      \put(3874,154){\makebox(0,0){\strut{}Augmentation weight $\lambda$}}%
      \put(3874,4709){\makebox(0,0){\strut{}Precision Oxford: LSH with SL-12}}%
    }%
    \gplbacktext
    \put(0,0){\includegraphics{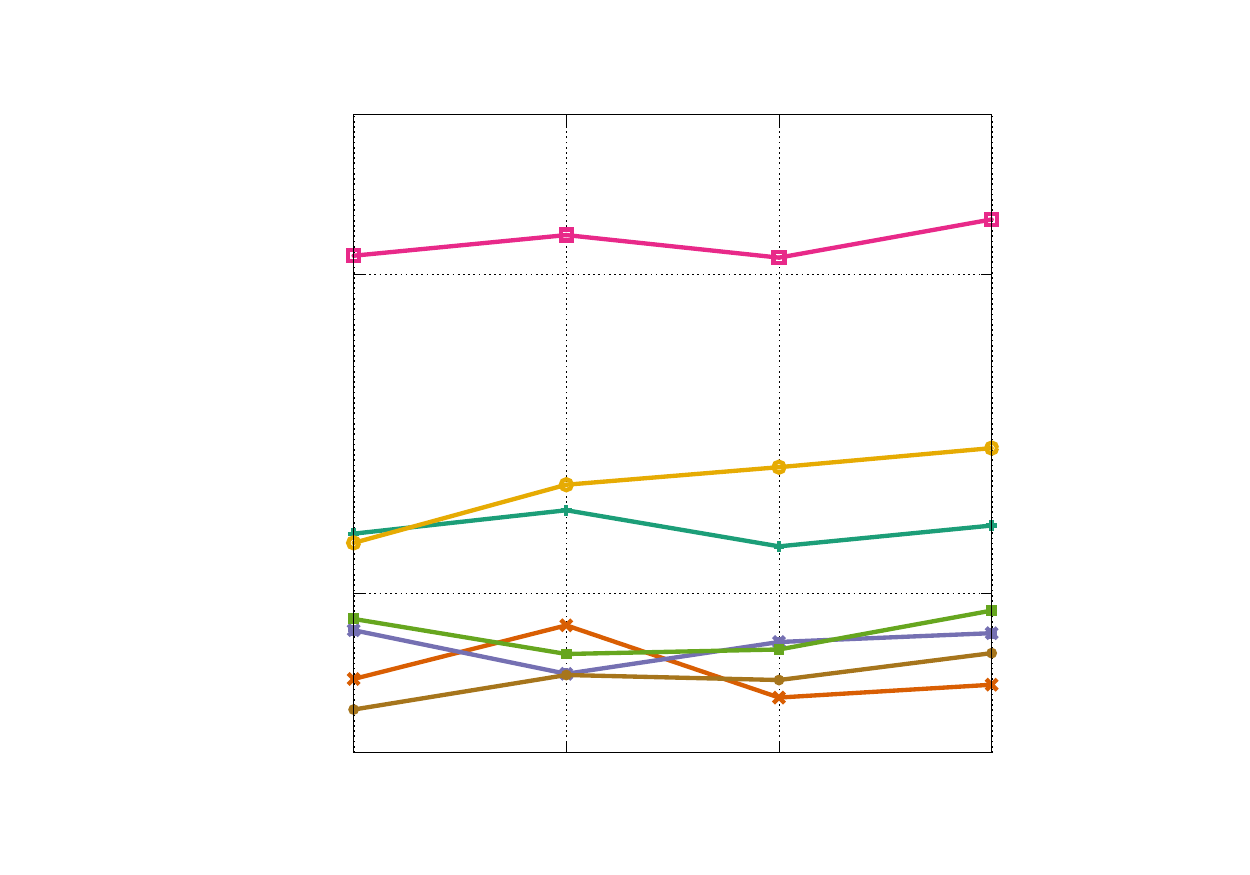}}%
    \gplfronttext
  \end{picture}
  }
  \adjustbox{width=0.45\columnwidth, trim=59pt 0pt 60pt 10pt, clip}{
  \begin{picture}(7200.00,5040.00)%
      \gdef\gplbacktext{}
    \gdef\gplfronttext{}
    \gplgaddtomacro\gplbacktext{%
      \csname LTb\endcsname%
      \put(1905,704){\makebox(0,0)[r]{\strut{}$0.1$}}%
      \csname LTb\endcsname%
      \put(1905,2804){\makebox(0,0)[r]{\strut{}$0.2$}}%
      \csname LTb\endcsname%
      \put(2037,484){\makebox(0,0){\strut{}0}}%
      \csname LTb\endcsname%
      \put(2650,484){\makebox(0,0){\strut{}4}}%
      \csname LTb\endcsname%
      \put(3262,484){\makebox(0,0){\strut{}8}}%
      \csname LTb\endcsname%
      \put(3875,484){\makebox(0,0){\strut{}16}}%
      \csname LTb\endcsname%
      \put(4487,484){\makebox(0,0){\strut{}32}}%
      \csname LTb\endcsname%
      \put(5100,484){\makebox(0,0){\strut{}64}}%
      \csname LTb\endcsname%
      \put(5712,484){\makebox(0,0){\strut{}128}}%
    }%
    \gplgaddtomacro\gplfronttext{%
      \csname LTb\endcsname%
      \put(1267,2541){\rotatebox{-270}{\makebox(0,0){\strut{}mAP}}}%
      \put(3874,154){\makebox(0,0){\strut{}Augmentation weight $\lambda$}}%
      \put(3874,4709){\makebox(0,0){\strut{}Precision Paris: BOF with SL-12}}%
    }%
    \gplbacktext
    \put(0,0){\includegraphics{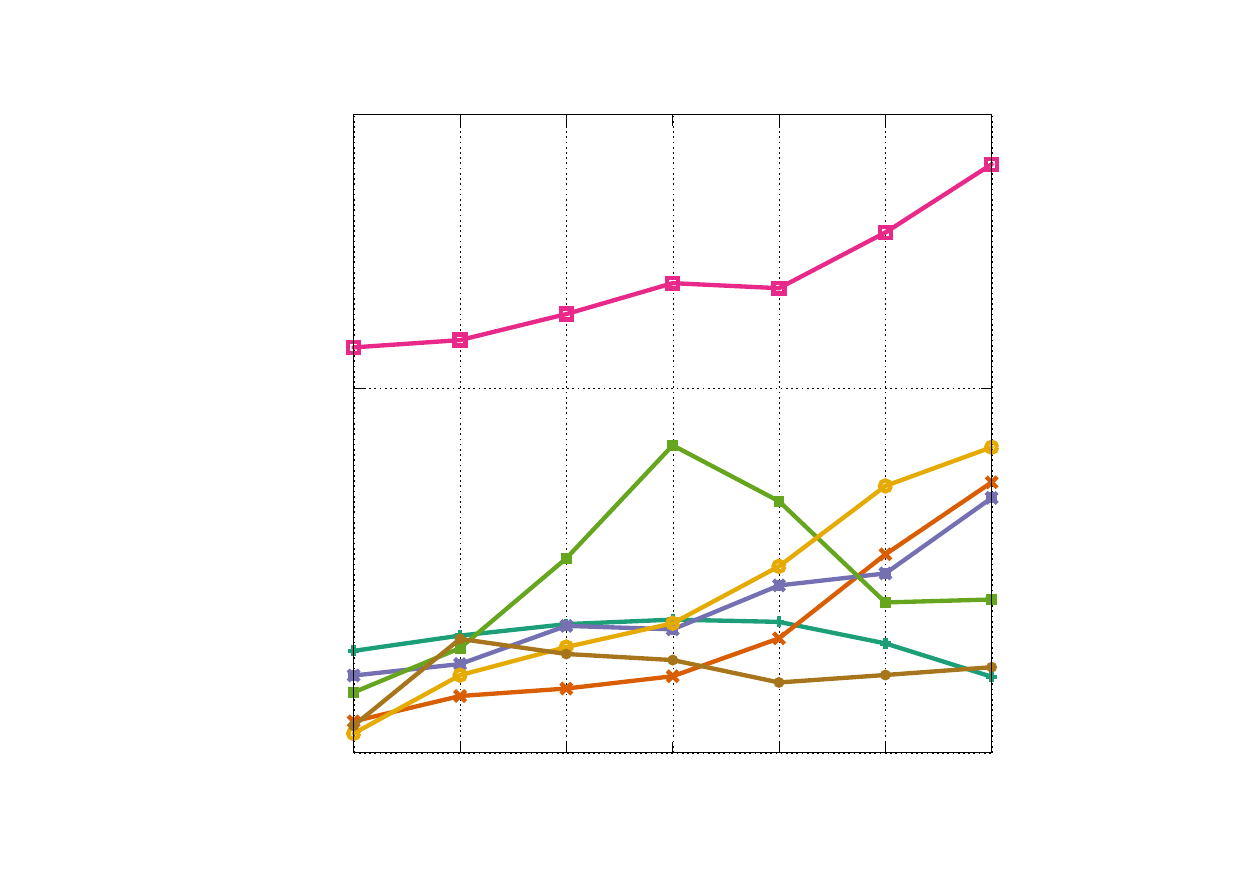}}%
    \gplfronttext
  \end{picture}
  }
  \adjustbox{width=0.45\columnwidth, trim=59pt 0pt 60pt 10pt, clip}{
  \begin{picture}(7200.00,5040.00)%
      \gdef\gplbacktext{}
    \gdef\gplfronttext{}
    \gplgaddtomacro\gplbacktext{%
      \csname LTb\endcsname%
      \put(1905,704){\makebox(0,0)[r]{\strut{}$0.2$}}%
      \csname LTb\endcsname%
      \put(1905,1623){\makebox(0,0)[r]{\strut{}$0.3$}}%
      \csname LTb\endcsname%
      \put(1905,2542){\makebox(0,0)[r]{\strut{}$0.4$}}%
      \csname LTb\endcsname%
      \put(1905,3460){\makebox(0,0)[r]{\strut{}$0.5$}}%
      \csname LTb\endcsname%
      \put(1905,4379){\makebox(0,0)[r]{\strut{}$0.6$}}%
      \csname LTb\endcsname%
      \put(2037,484){\makebox(0,0){\strut{}0}}%
      \csname LTb\endcsname%
      \put(2650,484){\makebox(0,0){\strut{}4}}%
      \csname LTb\endcsname%
      \put(3262,484){\makebox(0,0){\strut{}8}}%
      \csname LTb\endcsname%
      \put(3875,484){\makebox(0,0){\strut{}16}}%
      \csname LTb\endcsname%
      \put(4487,484){\makebox(0,0){\strut{}32}}%
      \csname LTb\endcsname%
      \put(5100,484){\makebox(0,0){\strut{}64}}%
      \csname LTb\endcsname%
      \put(5712,484){\makebox(0,0){\strut{}128}}%
    }%
    \gplgaddtomacro\gplfronttext{%
      \csname LTb\endcsname%
      \put(1267,2541){\rotatebox{-270}{\makebox(0,0){\strut{}mAP}}}%
      \put(3874,154){\makebox(0,0){\strut{}Augmentation weight $\lambda$}}%
      \put(3874,4709){\makebox(0,0){\strut{}Precision Holidays: BST with SL-12}}%
    }%
    \gplbacktext
    \put(0,0){\includegraphics{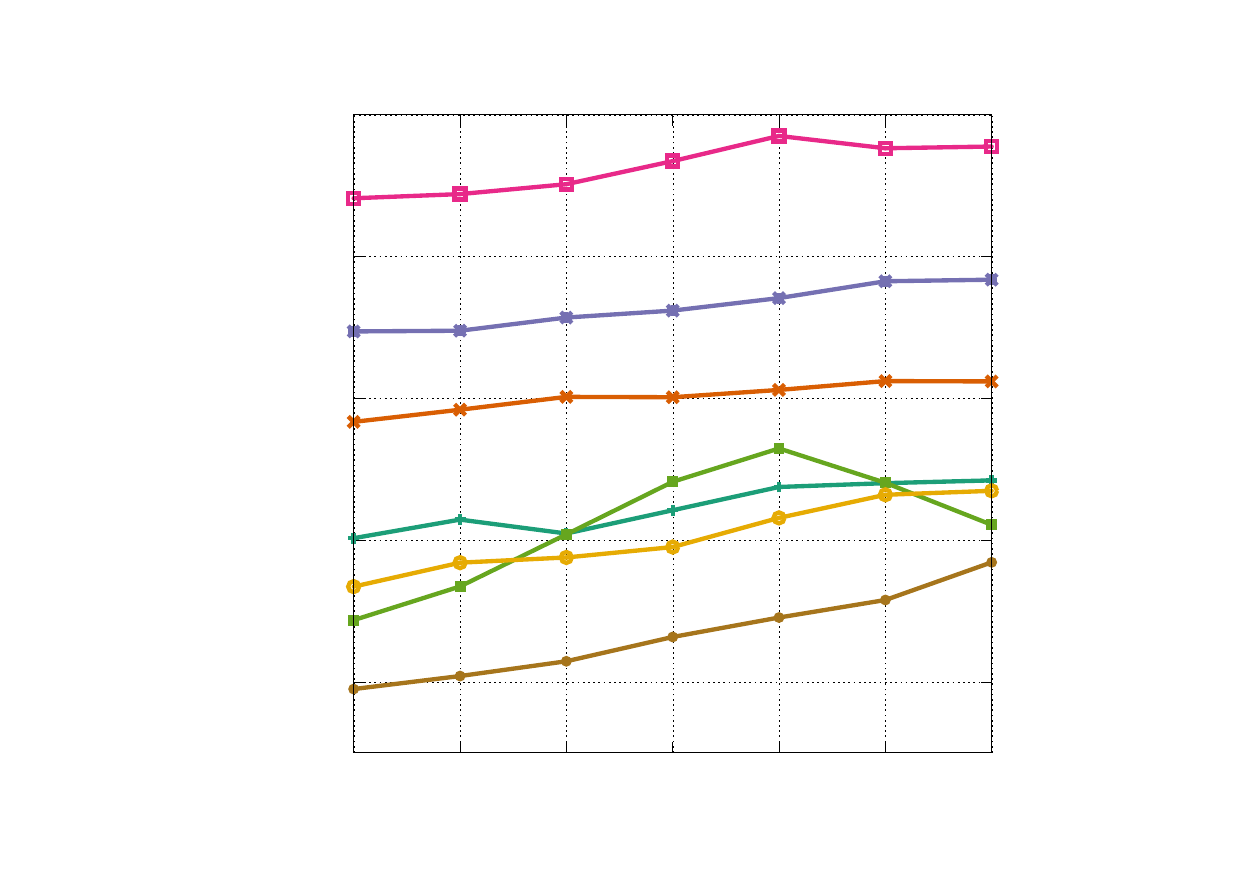}}%
    \gplfronttext
  \end{picture}
  }
  \adjustbox{width=0.2\columnwidth, trim=10pt 60pt 265pt 45pt, clip}{
  \begin{picture}(7200.00,5040.00)%
    \gdef\gplbacktext{}
    \gdef\gplfronttext{}
    \gplgaddtomacro\gplbacktext{%
    }%
    \gplgaddtomacro\gplfronttext{%
      \csname LTb\endcsname%
      \put(723,4011){\makebox(0,0)[l]{\strut{}BRIEF-256}}%
      \csname LTb\endcsname%
      \put(723,3791){\makebox(0,0)[l]{\strut{}ORB-256}}%
      \csname LTb\endcsname%
      \put(723,3571){\makebox(0,0)[l]{\strut{}BRISK-512}}%
      \csname LTb\endcsname%
      \put(723,3351){\makebox(0,0)[l]{\strut{}A-KAZE-486}}%
      \csname LTb\endcsname%
      \put(723,3131){\makebox(0,0)[l]{\strut{}FREAK-512}}%
      \csname LTb\endcsname%
      \put(723,2911){\makebox(0,0)[l]{\strut{}LDAHash-128}}%
      \csname LTb\endcsname%
      \put(723,2691){\makebox(0,0)[l]{\strut{}BinBoost-064}}%
    }%
    \gplbacktext
    \put(0,0){\includegraphics{plot-descriptor-key-eps-converted-to.pdf}}%
    \gplfronttext
  \end{picture} 
  }
  \caption{Results for \gls{sl}-12 with varying augmentation weight $\lambda$. A maximum of $1000$ descriptors has been computed for each image.}
  \label{fig:results-sl}
  \vspace{-15pt}
\end{figure*}

\subsection{Results}\label{sec:results}

We first review our results for the standard image retrieval datasets ZuBuD, Oxford, Paris and Holidays,
using a particular search method for each dataset.
In each case we computed the \gls{map}, using the public evaluation tool of~\cite{2007-philbin-object}.
In \figref{fig:results-kc} and \figref{fig:results-sl} we display the obtained \gls{map} scores when augmenting descriptors
with their keypoint coordinates (\gls{kc} with $I_u=I_v=8$) and semantic labels (\gls{sl}).
For each image, we computed a maximum of $1000$ descriptors,
sorted in descending order by their keypoints' response.

We evaluated the \gls{map} scores for increasing augmentation weights $\lambda$.
Note that for $\lambda=0$ (i.e. cue is not considered) the top and their respective bottom plots start from the same \gls{map} value.
We mention the original number of bits for each evaluated descriptor type with a suffix (e.g. BRIEF-$256$).

Two descriptors are considered to match if their Hamming distance $L_\mathcal{H}^\augmentationweight(\descriptor_\star, \descriptor_\star')$
lies within the threshold $\tau$.
For all datasets we set $\tau$ to 10\% of the augmented descriptors size (e.g. $\tau=27.2$ for BRIEF-$256$ with \gls{kc}-8x8).
Hence, $\tau$ grows as we increase $\augmentationweight$, reducing precision in case noise is added.
Since the \gls{sl} augmentation results in only distances of $1$ respectively $0$ although the encoding length is $12$ bits,
we adjusted our threshold $\tau$ accordingly and additionally evaluated higher $\lambda$ (i.e. $64$ and $128$ for \gls{bf} in \figref{fig:results-sl}).

\textbf{ZuBuD (\gls{bf})}:
The keypoint coordinates turn out to be a highly beneficial cue when combined with \gls{bf}.
The significant precision drop at $\augmentationweight=4$ for BinBoost, and at $\augmentationweight=32$ for LDAHash and ORB
marks the point at which the cues contribution $\bb$ saturates the descriptor $\descriptor$:
$L_\mathcal{H}^\augmentationweight(\descriptor_\star, \descriptor_\star') \approx \augmentationweight L_\mathcal{H}(\bb, \bb') \gg L_\mathcal{H}(\descriptor, \descriptor')$.
In this case, relevant image information stored in $\descriptor$ is neglected and the comparison is based only on the cues.
Clearly, $L_\mathcal{H}(\bb, \bb')$ is not sufficiently descriptive anymore to find true matches within thousands of images.
The smaller the used descriptor type, the earlier this happens.
This phenomena can be observed in all datasets.
Considering the semantic labels, the gain in precision is still observable yet more restrained.

\textbf{Oxford (\gls{lsh})}:
The precision gain for both cue types is mild, yet an improvement can be observed for $\lambda=1$
for every search method and descriptor type.

\textbf{Paris (\gls{bof})}:
In the Paris dataset we observe clear precision peaks for several descriptor types.
The \gls{bof} model profits from the additional information from our cues.

\textbf{Holidays (\gls{bst})}:
The employed \gls{bst} approach benefits from our cues, yet only marginally.
The Holidays dataset contains the most diverse images of all the datasets
(e.g. from buildings to nature to human faces).
In this challenging scenario our semantic labels are insufficient to provide significant additional information.

\begin{figure*}[ht!]
  \centering
  \adjustbox{width=0.65\columnwidth, trim=50pt 0pt 60pt 0pt, clip}{
\begin{picture}(7200.00,5040.00)%
    \gdef\gplbacktext{}
    \gdef\gplfronttext{}
    \gplgaddtomacro\gplbacktext{%
      \csname LTb\endcsname%
      \put(1223,484){\makebox(0,0){\strut{}$0$}}%
      \csname LTb\endcsname%
      \put(2142,484){\makebox(0,0){\strut{}$0.25$}}%
      \csname LTb\endcsname%
      \put(3061,484){\makebox(0,0){\strut{}$0.5$}}%
      \csname LTb\endcsname%
      \put(3979,484){\makebox(0,0){\strut{}$0.75$}}%
      \csname LTb\endcsname%
      \put(4898,484){\makebox(0,0){\strut{}$1$}}%
      \csname LTb\endcsname%
      \put(5030,704){\makebox(0,0)[l]{\strut{}$0$}}%
      \csname LTb\endcsname%
      \put(5030,1623){\makebox(0,0)[l]{\strut{}$0.25$}}%
      \csname LTb\endcsname%
      \put(5030,2542){\makebox(0,0)[l]{\strut{}$0.5$}}%
      \csname LTb\endcsname%
      \put(5030,3460){\makebox(0,0)[l]{\strut{}$0.75$}}%
      \csname LTb\endcsname%
      \put(5030,4379){\makebox(0,0)[l]{\strut{}$1$}}%
    }%
    \gplgaddtomacro\gplfronttext{%
      \csname LTb\endcsname%
      \put(5799,2541){\rotatebox{-270}{\makebox(0,0){\strut{}Precision (correct/reported associations)}}}%
      \put(3060,154){\makebox(0,0){\strut{}Recall (correct/possible associations)}}%
      \put(3060,4709){\makebox(0,0){\strut{}Precision-Recall: ORB-256 KC-8x8}}%
      \csname LTb\endcsname%
      \put(2543,2417){\makebox(0,0)[r]{\strut{}BF, $\lambda=0$}}%
      \csname LTb\endcsname%
      \put(2543,2197){\makebox(0,0)[r]{\strut{}BF, $\lambda=16$}}%
      \csname LTb\endcsname%
      \put(2543,1977){\makebox(0,0)[r]{\strut{}LSH, $\lambda=0$}}%
      \csname LTb\endcsname%
      \put(2543,1757){\makebox(0,0)[r]{\strut{}LSH, $\lambda=16$}}%
      \csname LTb\endcsname%
      \put(2543,1537){\makebox(0,0)[r]{\strut{}BOF, $\lambda=0$}}%
      \csname LTb\endcsname%
      \put(2543,1317){\makebox(0,0)[r]{\strut{}BOF, $\lambda=16$}}%
      \csname LTb\endcsname%
      \put(2543,1097){\makebox(0,0)[r]{\strut{}BST, $\lambda=0$}}%
      \csname LTb\endcsname%
      \put(2543,877){\makebox(0,0)[r]{\strut{}BST, $\lambda=16$}}%
    }%
    \gplbacktext
    \put(0,0){\includegraphics{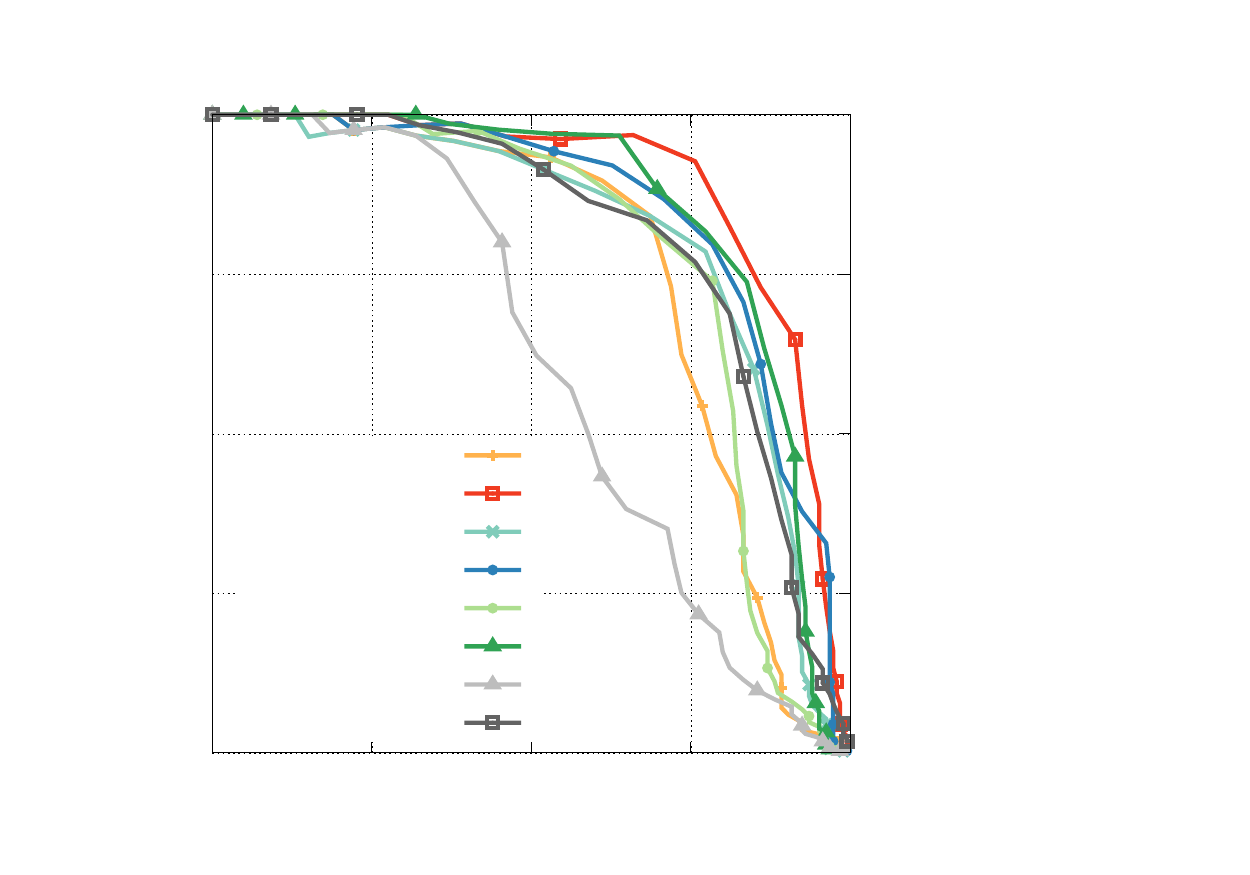}}%
    \gplfronttext
  \end{picture}
}
\adjustbox{width=0.65\columnwidth, trim=50pt 0pt 60pt 0pt, clip}{
\begin{picture}(7200.00,5040.00)%
    \gdef\gplbacktext{}
    \gdef\gplfronttext{}
    \gplgaddtomacro\gplbacktext{%
      \csname LTb\endcsname%
      \put(1223,484){\makebox(0,0){\strut{}$0.5$}}%
      \csname LTb\endcsname%
      \put(3061,484){\makebox(0,0){\strut{}$0.75$}}%
      \csname LTb\endcsname%
      \put(4898,484){\makebox(0,0){\strut{}$1$}}%
      \csname LTb\endcsname%
      \put(5030,704){\makebox(0,0)[l]{\strut{}$0$}}%
      \csname LTb\endcsname%
      \put(5030,1623){\makebox(0,0)[l]{\strut{}$0.25$}}%
      \csname LTb\endcsname%
      \put(5030,2542){\makebox(0,0)[l]{\strut{}$0.5$}}%
      \csname LTb\endcsname%
      \put(5030,3460){\makebox(0,0)[l]{\strut{}$0.75$}}%
      \csname LTb\endcsname%
      \put(5030,4379){\makebox(0,0)[l]{\strut{}$1$}}%
    }%
    \gplgaddtomacro\gplfronttext{%
      \csname LTb\endcsname%
      \put(5799,2541){\rotatebox{-270}{\makebox(0,0){\strut{}Precision (correct/reported associations)}}}%
      \put(3060,154){\makebox(0,0){\strut{}Recall (correct/possible associations)}}%
      \put(3060,4709){\makebox(0,0){\strut{}Precision-Recall: BRISK-512 KC-8x8}}%
      \csname LTb\endcsname%
      \put(2543,2417){\makebox(0,0)[r]{\strut{}BF, $\lambda=0$}}%
      \csname LTb\endcsname%
      \put(2543,2197){\makebox(0,0)[r]{\strut{}BF, $\lambda=16$}}%
      \csname LTb\endcsname%
      \put(2543,1977){\makebox(0,0)[r]{\strut{}LSH, $\lambda=0$}}%
      \csname LTb\endcsname%
      \put(2543,1757){\makebox(0,0)[r]{\strut{}LSH, $\lambda=16$}}%
      \csname LTb\endcsname%
      \put(2543,1537){\makebox(0,0)[r]{\strut{}BOF, $\lambda=0$}}%
      \csname LTb\endcsname%
      \put(2543,1317){\makebox(0,0)[r]{\strut{}BOF, $\lambda=16$}}%
      \csname LTb\endcsname%
      \put(2543,1097){\makebox(0,0)[r]{\strut{}BST, $\lambda=0$}}%
      \csname LTb\endcsname%
      \put(2543,877){\makebox(0,0)[r]{\strut{}BST, $\lambda=16$}}%
    }%
    \gplbacktext
    \put(0,0){\includegraphics{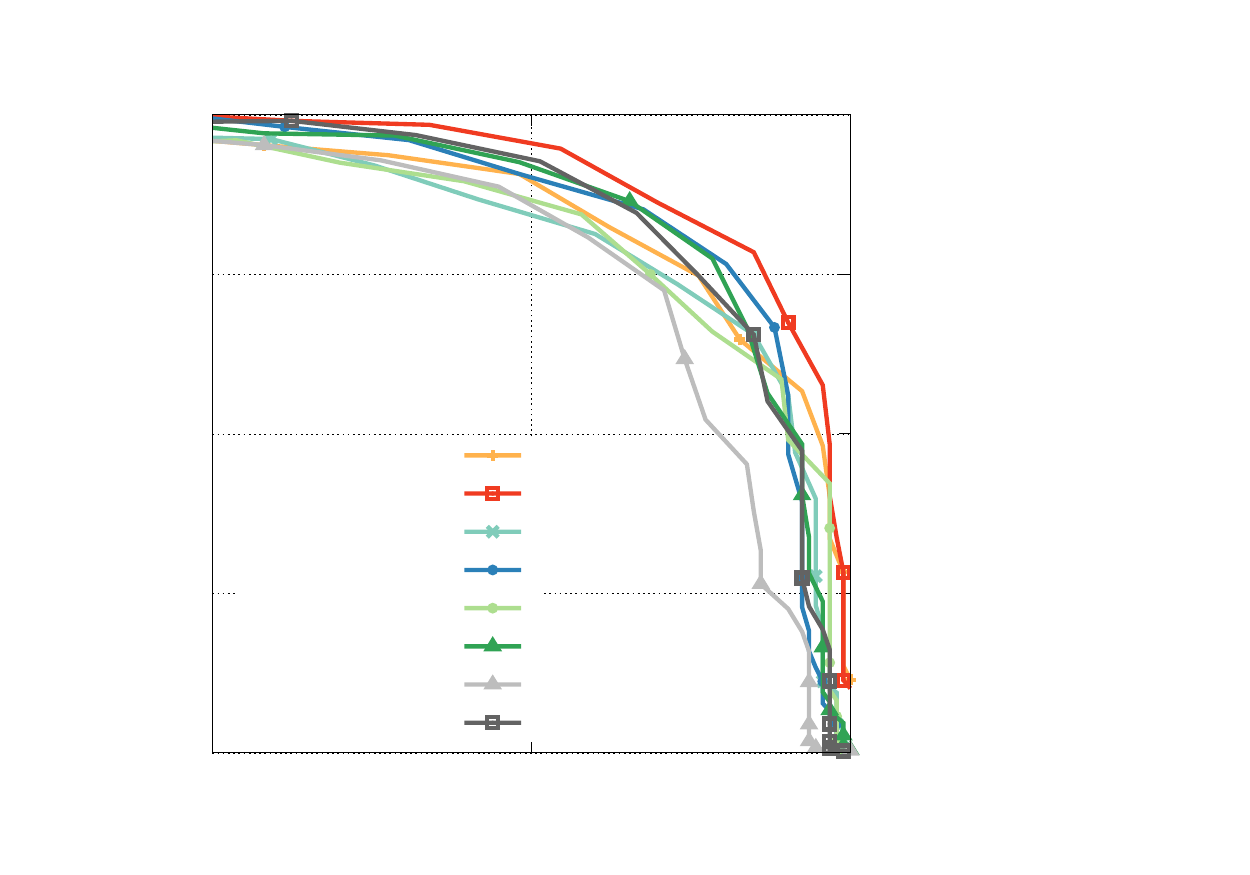}}%
    \gplfronttext
  \end{picture} 
}
\adjustbox{width=0.65\columnwidth, trim=50pt 0pt 60pt 0pt, clip}{
\begin{picture}(7200.00,5040.00)%
    \gdef\gplbacktext{}
    \gdef\gplfronttext{}
    \gplgaddtomacro\gplbacktext{%
      \csname LTb\endcsname%
      \put(1223,484){\makebox(0,0){\strut{}$0.5$}}%
      \csname LTb\endcsname%
      \put(3061,484){\makebox(0,0){\strut{}$0.75$}}%
      \csname LTb\endcsname%
      \put(4898,484){\makebox(0,0){\strut{}$1$}}%
      \csname LTb\endcsname%
      \put(5030,704){\makebox(0,0)[l]{\strut{}$0$}}%
      \csname LTb\endcsname%
      \put(5030,1623){\makebox(0,0)[l]{\strut{}$0.25$}}%
      \csname LTb\endcsname%
      \put(5030,2542){\makebox(0,0)[l]{\strut{}$0.5$}}%
      \csname LTb\endcsname%
      \put(5030,3460){\makebox(0,0)[l]{\strut{}$0.75$}}%
      \csname LTb\endcsname%
      \put(5030,4379){\makebox(0,0)[l]{\strut{}$1$}}%
    }%
    \gplgaddtomacro\gplfronttext{%
      \csname LTb\endcsname%
      \put(5799,2541){\rotatebox{-270}{\makebox(0,0){\strut{}Precision (correct/reported associations)}}}%
      \put(3060,154){\makebox(0,0){\strut{}Recall (correct/possible associations)}}%
      \put(3060,4709){\makebox(0,0){\strut{}Precision-Recall: FREAK-512 KC-8x8}}%
      \csname LTb\endcsname%
      \put(2543,2417){\makebox(0,0)[r]{\strut{}BF, $\lambda=0$}}%
      \csname LTb\endcsname%
      \put(2543,2197){\makebox(0,0)[r]{\strut{}BF, $\lambda=16$}}%
      \csname LTb\endcsname%
      \put(2543,1977){\makebox(0,0)[r]{\strut{}LSH, $\lambda=0$}}%
      \csname LTb\endcsname%
      \put(2543,1757){\makebox(0,0)[r]{\strut{}LSH, $\lambda=16$}}%
      \csname LTb\endcsname%
      \put(2543,1537){\makebox(0,0)[r]{\strut{}BOF, $\lambda=0$}}%
      \csname LTb\endcsname%
      \put(2543,1317){\makebox(0,0)[r]{\strut{}BOF, $\lambda=16$}}%
      \csname LTb\endcsname%
      \put(2543,1097){\makebox(0,0)[r]{\strut{}BST, $\lambda=0$}}%
      \csname LTb\endcsname%
      \put(2543,877){\makebox(0,0)[r]{\strut{}BST, $\lambda=16$}}%
    }%
    \gplbacktext
    \put(0,0){\includegraphics{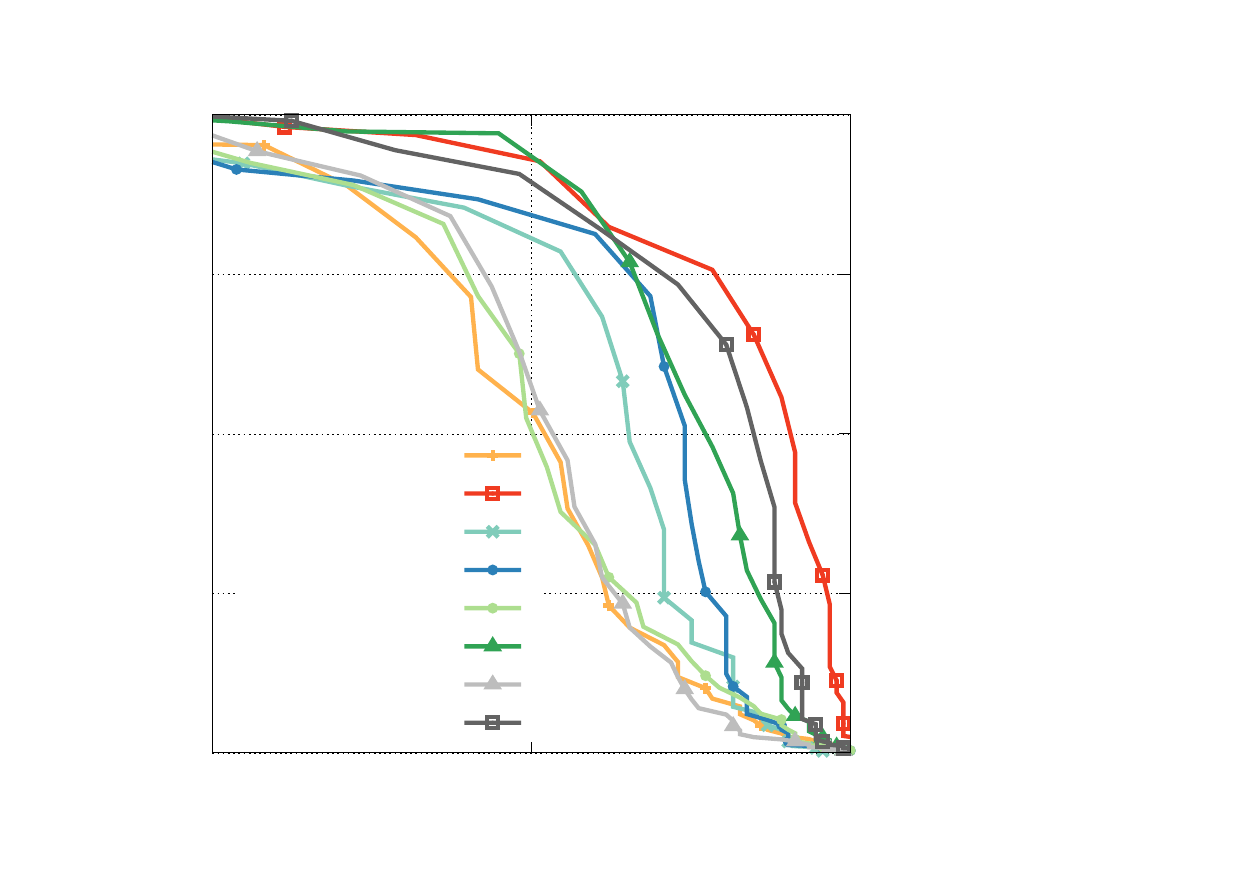}}%
    \gplfronttext
  \end{picture} 
}
\adjustbox{width=0.65\columnwidth, trim=50pt 0pt 60pt 0pt, clip}{
\begin{picture}(7200.00,5040.00)%
    \gdef\gplbacktext{}
    \gdef\gplfronttext{}
    \gplgaddtomacro\gplbacktext{%
      \csname LTb\endcsname%
      \put(1223,484){\makebox(0,0){\strut{}$0.5$}}%
      \csname LTb\endcsname%
      \put(3061,484){\makebox(0,0){\strut{}$0.75$}}%
      \csname LTb\endcsname%
      \put(4898,484){\makebox(0,0){\strut{}$1$}}%
      \csname LTb\endcsname%
      \put(5030,704){\makebox(0,0)[l]{\strut{}$0$}}%
      \csname LTb\endcsname%
      \put(5030,1623){\makebox(0,0)[l]{\strut{}$0.25$}}%
      \csname LTb\endcsname%
      \put(5030,2542){\makebox(0,0)[l]{\strut{}$0.5$}}%
      \csname LTb\endcsname%
      \put(5030,3460){\makebox(0,0)[l]{\strut{}$0.75$}}%
      \csname LTb\endcsname%
      \put(5030,4379){\makebox(0,0)[l]{\strut{}$1$}}%
    }%
    \gplgaddtomacro\gplfronttext{%
      \csname LTb\endcsname%
      \put(5799,2541){\rotatebox{-270}{\makebox(0,0){\strut{}Precision (correct/reported associations)}}}%
      \put(3060,154){\makebox(0,0){\strut{}Recall (correct/possible associations)}}%
      \put(3060,4709){\makebox(0,0){\strut{}Precision-Recall: A-KAZE-486 KC-8x8}}%
      \csname LTb\endcsname%
      \put(2543,2417){\makebox(0,0)[r]{\strut{}BF, $\lambda=0$}}%
      \csname LTb\endcsname%
      \put(2543,2197){\makebox(0,0)[r]{\strut{}BF, $\lambda=16$}}%
      \csname LTb\endcsname%
      \put(2543,1977){\makebox(0,0)[r]{\strut{}LSH, $\lambda=0$}}%
      \csname LTb\endcsname%
      \put(2543,1757){\makebox(0,0)[r]{\strut{}LSH, $\lambda=16$}}%
      \csname LTb\endcsname%
      \put(2543,1537){\makebox(0,0)[r]{\strut{}BOF, $\lambda=0$}}%
      \csname LTb\endcsname%
      \put(2543,1317){\makebox(0,0)[r]{\strut{}BOF, $\lambda=16$}}%
      \csname LTb\endcsname%
      \put(2543,1097){\makebox(0,0)[r]{\strut{}BST, $\lambda=0$}}%
      \csname LTb\endcsname%
      \put(2543,877){\makebox(0,0)[r]{\strut{}BST, $\lambda=16$}}%
    }%
    \gplbacktext
    \put(0,0){\includegraphics{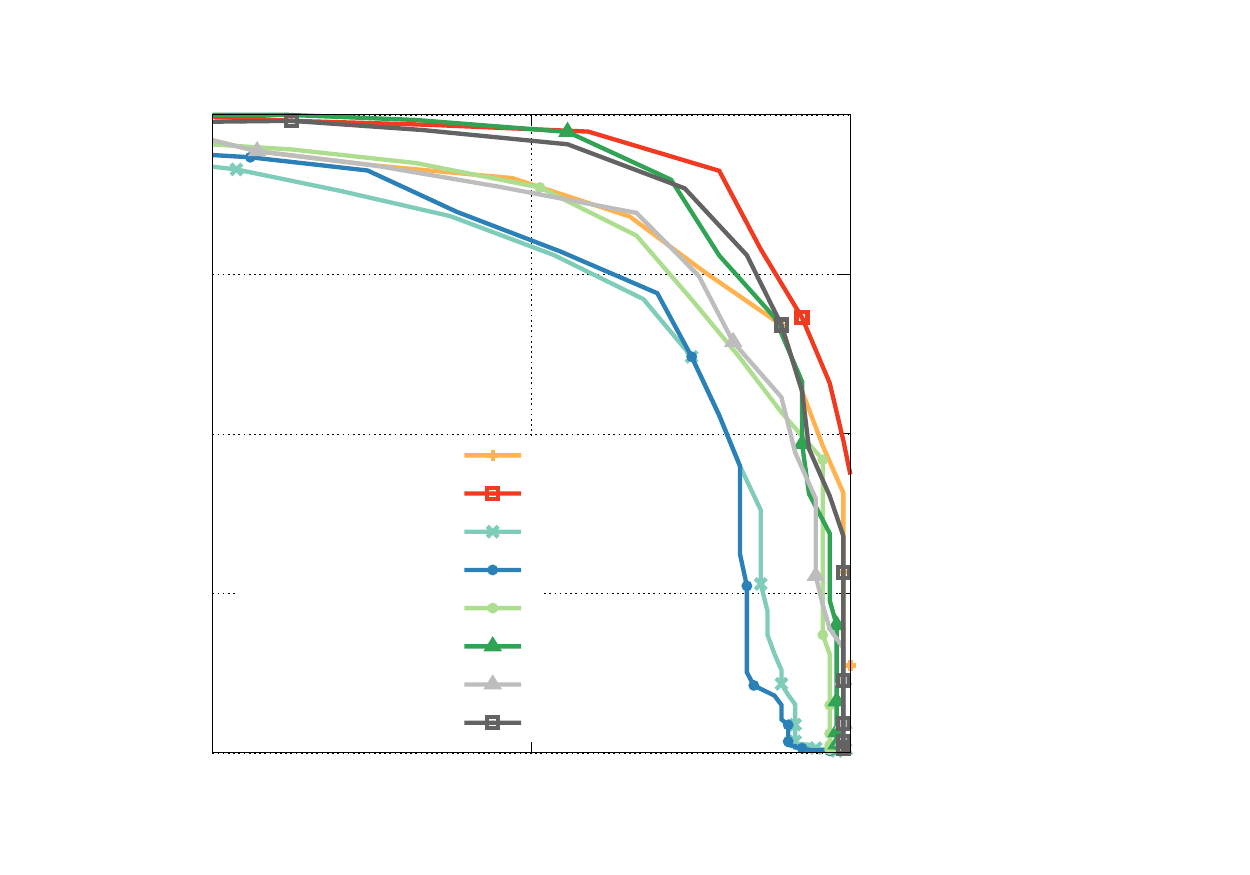}}%
    \gplfronttext
  \end{picture}
}
\adjustbox{width=0.65\columnwidth, trim=50pt 0pt 60pt 0pt, clip}{
\begin{picture}(7200.00,5040.00)%
    \gdef\gplbacktext{}
    \gdef\gplfronttext{}
    \gplgaddtomacro\gplbacktext{%
      \csname LTb\endcsname%
      \put(1223,484){\makebox(0,0){\strut{}$0.5$}}%
      \csname LTb\endcsname%
      \put(3061,484){\makebox(0,0){\strut{}$0.75$}}%
      \csname LTb\endcsname%
      \put(4898,484){\makebox(0,0){\strut{}$1$}}%
      \csname LTb\endcsname%
      \put(5030,704){\makebox(0,0)[l]{\strut{}$0$}}%
      \csname LTb\endcsname%
      \put(5030,1623){\makebox(0,0)[l]{\strut{}$0.25$}}%
      \csname LTb\endcsname%
      \put(5030,2542){\makebox(0,0)[l]{\strut{}$0.5$}}%
      \csname LTb\endcsname%
      \put(5030,3460){\makebox(0,0)[l]{\strut{}$0.75$}}%
      \csname LTb\endcsname%
      \put(5030,4379){\makebox(0,0)[l]{\strut{}$1$}}%
    }%
    \gplgaddtomacro\gplfronttext{%
      \csname LTb\endcsname%
      \put(5799,2541){\rotatebox{-270}{\makebox(0,0){\strut{}Precision (correct/reported associations)}}}%
      \put(3060,154){\makebox(0,0){\strut{}Recall (correct/possible associations)}}%
      \put(3060,4709){\makebox(0,0){\strut{}Precision-Recall: LDAHash-128 KC-8x8}}%
      \csname LTb\endcsname%
      \put(2543,2417){\makebox(0,0)[r]{\strut{}BF, $\lambda=0$}}%
      \csname LTb\endcsname%
      \put(2543,2197){\makebox(0,0)[r]{\strut{}BF, $\lambda=16$}}%
      \csname LTb\endcsname%
      \put(2543,1977){\makebox(0,0)[r]{\strut{}LSH, $\lambda=0$}}%
      \csname LTb\endcsname%
      \put(2543,1757){\makebox(0,0)[r]{\strut{}LSH, $\lambda=16$}}%
      \csname LTb\endcsname%
      \put(2543,1537){\makebox(0,0)[r]{\strut{}BOF, $\lambda=0$}}%
      \csname LTb\endcsname%
      \put(2543,1317){\makebox(0,0)[r]{\strut{}BOF, $\lambda=16$}}%
      \csname LTb\endcsname%
      \put(2543,1097){\makebox(0,0)[r]{\strut{}BST, $\lambda=0$}}%
      \csname LTb\endcsname%
      \put(2543,877){\makebox(0,0)[r]{\strut{}BST, $\lambda=16$}}%
    }%
    \gplbacktext
    \put(0,0){\includegraphics{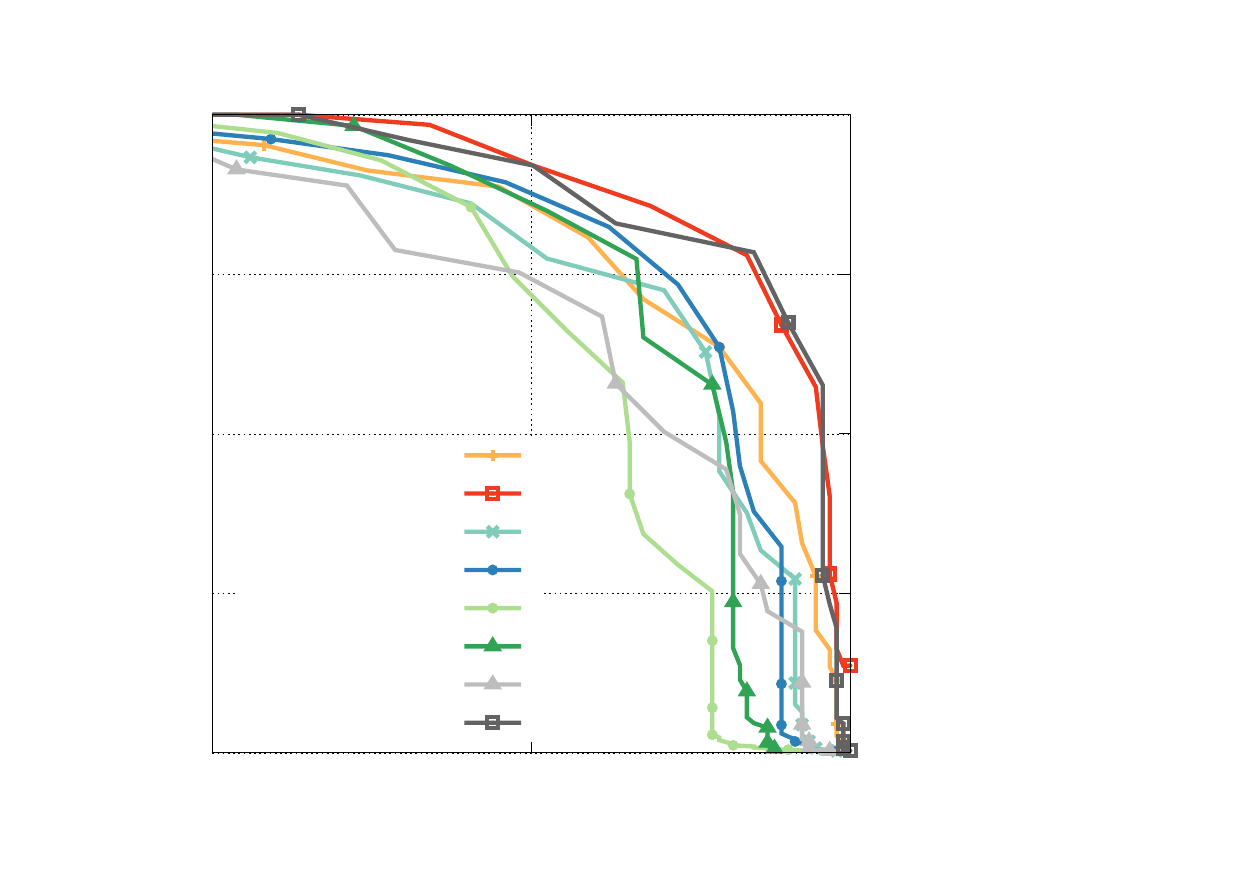}}%
    \gplfronttext
  \end{picture}
}
\adjustbox{width=0.65\columnwidth, trim=50pt 0pt 60pt 0pt, clip}{
\begin{picture}(7200.00,5040.00)%
    \gdef\gplbacktext{}
    \gdef\gplfronttext{}
    \gplgaddtomacro\gplbacktext{%
      \csname LTb\endcsname%
      \put(1223,484){\makebox(0,0){\strut{}$0$}}%
      \csname LTb\endcsname%
      \put(2142,484){\makebox(0,0){\strut{}$0.25$}}%
      \csname LTb\endcsname%
      \put(3061,484){\makebox(0,0){\strut{}$0.5$}}%
      \csname LTb\endcsname%
      \put(3979,484){\makebox(0,0){\strut{}$0.75$}}%
      \csname LTb\endcsname%
      \put(4898,484){\makebox(0,0){\strut{}$1$}}%
      \csname LTb\endcsname%
      \put(5030,704){\makebox(0,0)[l]{\strut{}$0$}}%
      \csname LTb\endcsname%
      \put(5030,1623){\makebox(0,0)[l]{\strut{}$0.25$}}%
      \csname LTb\endcsname%
      \put(5030,2542){\makebox(0,0)[l]{\strut{}$0.5$}}%
      \csname LTb\endcsname%
      \put(5030,3460){\makebox(0,0)[l]{\strut{}$0.75$}}%
      \csname LTb\endcsname%
      \put(5030,4379){\makebox(0,0)[l]{\strut{}$1$}}%
    }%
    \gplgaddtomacro\gplfronttext{%
      \csname LTb\endcsname%
      \put(5799,2541){\rotatebox{-270}{\makebox(0,0){\strut{}Precision (correct/reported associations)}}}%
      \put(3060,154){\makebox(0,0){\strut{}Recall (correct/possible associations)}}%
      \put(3060,4709){\makebox(0,0){\strut{}Precision-Recall: BinBoost-064 KC-8x8}}%
      \csname LTb\endcsname%
      \put(2543,2417){\makebox(0,0)[r]{\strut{}BF, $\lambda=0$}}%
      \csname LTb\endcsname%
      \put(2543,2197){\makebox(0,0)[r]{\strut{}BF, $\lambda=16$}}%
      \csname LTb\endcsname%
      \put(2543,1977){\makebox(0,0)[r]{\strut{}LSH, $\lambda=0$}}%
      \csname LTb\endcsname%
      \put(2543,1757){\makebox(0,0)[r]{\strut{}LSH, $\lambda=16$}}%
      \csname LTb\endcsname%
      \put(2543,1537){\makebox(0,0)[r]{\strut{}BOF, $\lambda=0$}}%
      \csname LTb\endcsname%
      \put(2543,1317){\makebox(0,0)[r]{\strut{}BOF, $\lambda=16$}}%
      \csname LTb\endcsname%
      \put(2543,1097){\makebox(0,0)[r]{\strut{}BST, $\lambda=0$}}%
      \csname LTb\endcsname%
      \put(2543,877){\makebox(0,0)[r]{\strut{}BST, $\lambda=16$}}%
    }%
    \gplbacktext
    \put(0,0){\includegraphics{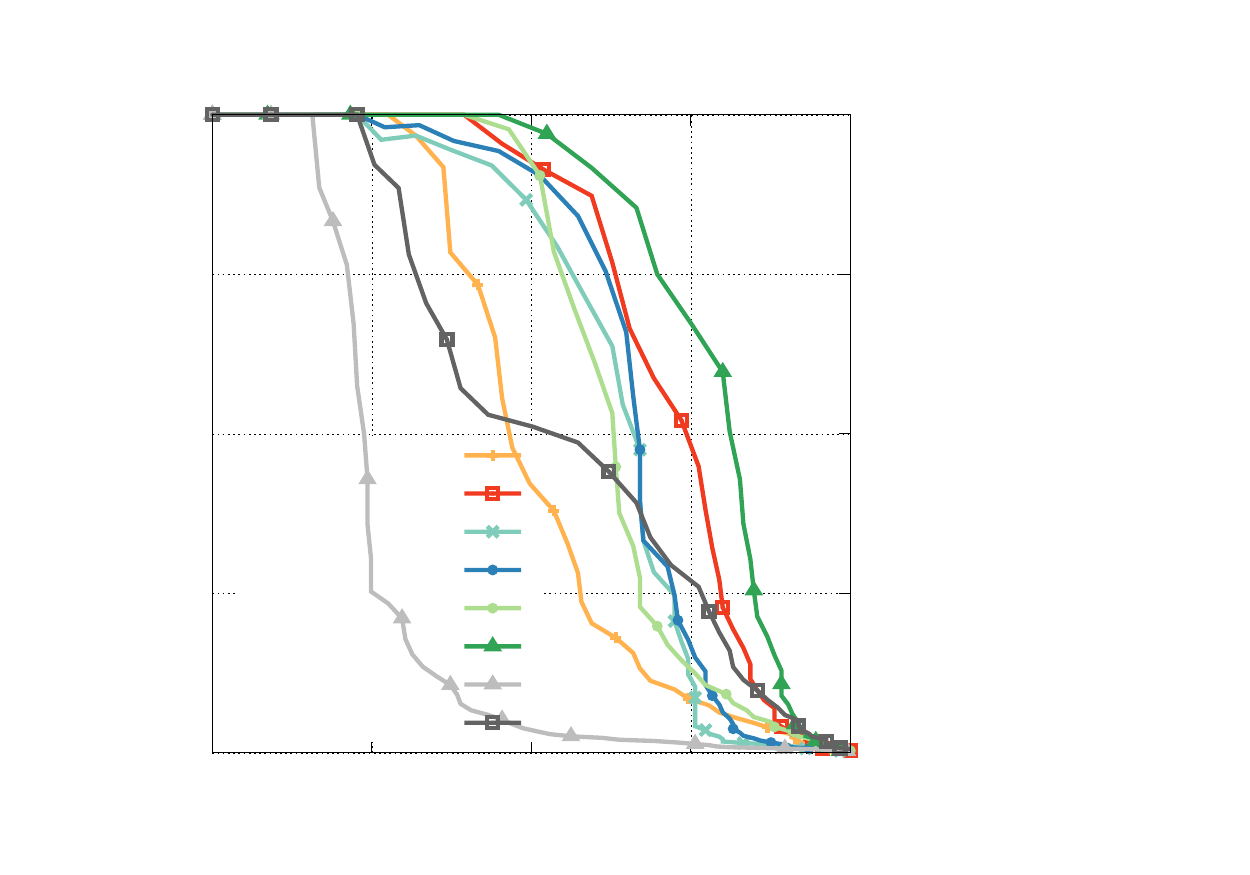}}%
    \gplfronttext
  \end{picture}
}
  \caption{\gls{pr} analysis on KITTI sequence 00 for various descriptor types for the continuous example cue \gls{kc}-8x8.
    For all search methods and descriptors we evaluated the cue weighting $\augmentationweight=0$ (cue ignored) and $\augmentationweight=16$.
    (see \figref{fig:motivation} for the \gls{pr} curves of BRIEF).}
  \label{fig:results-kitti}
  \vspace{-15pt}
\end{figure*}
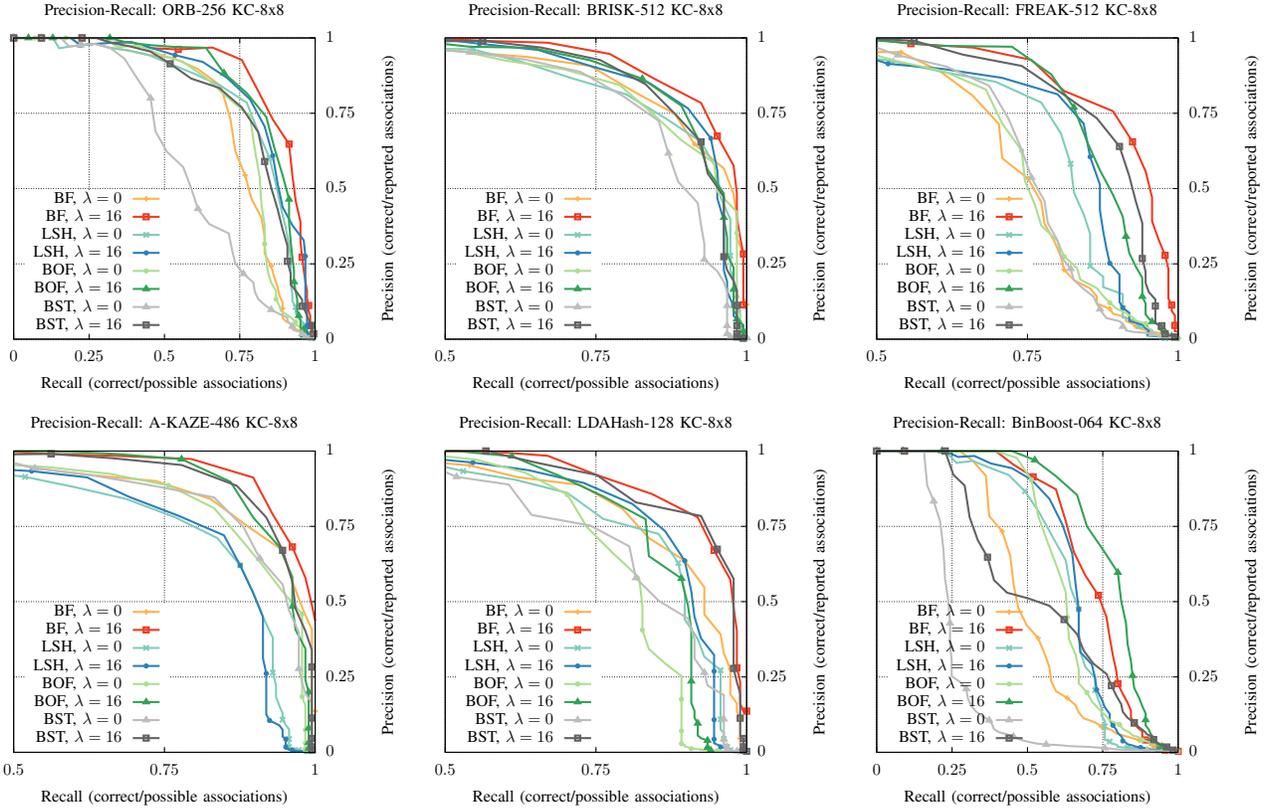

\textbf{KITTI}:
For the evaluation on the KITTI dataset we built the image database incrementally for each approach.
While processing the sequence of images, every image was first queried against the database of previous images
and then added in a subsequent step, followed by a training phase.
Clearly, this increases the computational demand for all methods considered as the database cannot be constructed in a single step.
We processed every 10th image in a sequence, to avoid capturing many noisy descriptor versions.

In contrast to the previous datasets (\figref{fig:results-kc} and \figref{fig:results-sl}), there are no rotated images in KITTI.
Hence when revisiting a place, similar features are automatically detected in similar image regions.
This clearly increases the utility of our keypoint coordinates cue, which is observable in our results.

Instead of the \gls{map} we computed the complete \gls{pr} curves for KITTI sequence 00.
In \figref{fig:motivation} we display the resulting performance boost obtained by augmenting BRIEF-256 descriptors
with our continuous and our selector cue.
In \figref{fig:results-kitti} we display \gls{pr} curves for $\augmentationweight=0$ respectively
$\augmentationweight=16$ in comparison with each search method for the remaining 6 descriptor types.

As in the previous datasets, a clear improvement of the search precision is observable.
The improvement however, is significantly more pronounced in KITTI as in the previous datasets.
This is due to the mentioned missing rotated imagery and road bound image acquisition.
Impressively, the evaluated \gls{bst} approach (black) manages to reach almost \gls{bf} (red) precision (FREAK, A-KAZE, LDAHash) when using our cues.
The semantic cues do not manage to improve the performance as much as the keypoint coordinate cues in this case (\figref{fig:motivation}).
Yet, they have a positive effect on precision that is consistent with the results of the previous datasets.

In \tabref{tab:runtime} we list the measured mean query processing times of the approaches
considered in \figref{fig:motivation} and \figref{fig:results-kitti} on KITTI.
We separately examined the specific augmentation weights $\augmentationweight\in\{0,1,16,32\}$
for the continuous augmentation case (\gls{kc}).

\begin{table}[h!]
  \centering
  \vspace{-5pt}
  \begin{subtable}{0.55\columnwidth}
    \centering\tiny
    Augmentation weight $\augmentationweight=0$\vspace{2pt}\\
    \begin{tabular}{|l|c|c|c|c|c|c|c|c|c|}
    \hline
    & \gls{bf} & \gls{lsh} & \gls{bof} & \gls{bst} \\
    \hline
    BRIEF-256 & 2.032 & 0.950 & 0.153 & 0.002 \\
    ORB-256 & 2.689 & 1.133 & 0.170 & 0.001 \\
    BRISK-512 & 3.781 & 1.139 & 0.261 & 0.002 \\
    A-KAZE-486 & 3.517 & 1.078 & 0.259 & 0.002 \\
    FREAK-512 & 3.023 & 1.010 & 0.262 & 0.003 \\
    LDAHash-128 & 2.561 & 1.319 & 0.090 & 0.001 \\
    BinBoost-064 & 2.693 & 0.735 & 0.084 & 0.002 \\
    \hline
    \end{tabular}
  \end{subtable}
  \begin{subtable}{0.4\columnwidth}
    \centering\tiny
    Augmentation weight $\augmentationweight=1$\vspace{2pt}\\
    \begin{tabular}{|c|c|c|c|c|c|c|c|c|}
    \hline
    \gls{bf} & \gls{lsh} & \gls{bof} & \gls{bst} \\
    \hline
    3.141 & 1.015 & 0.207 & 0.002 \\
    3.046 & 1.087 & 0.190 & 0.002 \\
    3.430 & 1.208 & 0.265 & 0.002 \\
    3.711 & 1.096 & 0.281 & 0.002 \\
    3.358 & 0.992 & 0.267 & 0.003 \\
    2.789 & 1.195 & 0.087 & 0.002 \\
    3.035 & 0.727 & 0.080 & 0.007 \\
    \hline
    \end{tabular}
  \end{subtable}\vspace{5pt}\\
  \begin{subtable}{0.55\columnwidth}
    \centering\tiny
    Augmentation weight $\augmentationweight=16$\vspace{2pt}\\
    \begin{tabular}{|l|c|c|c|c|c|c|c|c|c|}
    \hline
    BRIEF-256 & 3.251 & 0.907 & 0.215 & 0.003 \\
    ORB-256 & 3.858 & 1.081 & 0.202 & 0.002 \\
    BRISK-512 & 3.296 & 0.911 & 0.313 & 0.004 \\
    A-KAZE-486 & 3.853 & 0.943 & 0.306 & 0.003 \\
    FREAK-512 & 3.146 & 0.824 & 0.354 & 0.006 \\
    LDAHash-128 & 2.802 & 0.711 & 0.112 & 0.002 \\
    BinBoost-064 & 3.005 & 1.280 & 0.086 & 0.009 \\ 
    \hline
    \end{tabular}
  \end{subtable}
  \begin{subtable}{0.4\columnwidth}
    \centering\tiny
    Augmentation weight $\augmentationweight=32$\vspace{2pt}\\
    \begin{tabular}{|c|c|c|c|c|c|c|c|c|}
    \hline
    4.328 & 0.918 & 0.263 & 0.005 \\
    4.199 & 0.829 & 0.280 & 0.008 \\
    4.040 & 0.929 & 0.365 & 0.011 \\
    4.114 & 0.779 & 0.375 & 0.009 \\
    3.749 & 0.884 & 0.433 & 0.011 \\
    3.025 & 0.723 & 0.133 & 0.008 \\
    3.273 & 2.917 & 0.111 & 0.009 \\
    \hline
    \end{tabular}
  \end{subtable}
  \caption{Mean processing times $\overline{t}$ (seconds) for differently weighted \gls{kc} augmentations on KITTI sequence 00.
  We report the average $\overline{t}$ over 10 runs.}
  \label{tab:runtime}
  \vspace{5pt}
\end{table}

In the case of $\augmentationweight=0$, the augmentation is not considered
and hence the descriptor size is minimal.
As expected, the processing time $\overline{t}$ is the smallest compared to $\augmentationweight>0$
for most approaches.
For $\augmentationweight=16$ and $\augmentationweight=32$ the processing times are still adequate
for most search approaches.

\section{Conclusions}\label{sec:conclusions}

In this paper, we presented approach that 
improves the precision for feature-based \gls{vpr} by embedding continuous
and selector cues into binary feature descriptors.
Our approach is purely supplementary to the descriptor computation
and the similarity search method.
We implemented and evaluated our approach on several standard benchmark datasets
and covered a vast number of state-of-the-art binary descriptor type and search method combinations.
Our results suggest that our strategy is effective and increases \gls{vpr} precision,
regardless of the descriptor type and search method.

\cleardoublepage
\bibliographystyle{ieeetr}
\bibliography{robots}

\end{document}